\documentclass[preprint,12pt]{elsarticle}

\usepackage{amssymb}
\usepackage{amsmath}
\usepackage{graphicx}%
\usepackage{multirow}%
\usepackage{amsmath,amssymb,amsfonts}%
\usepackage{amsthm}%
\usepackage{mathrsfs}%
\usepackage[title]{appendix}%
\usepackage{xcolor}%
\usepackage{textcomp}%
\usepackage{manyfoot}%
\usepackage{booktabs}%
\usepackage{algorithm}%
\usepackage{algorithmicx}%
\usepackage{algpseudocode}%
\usepackage{listings}%

\usepackage{subcaption}
\usepackage{tabularx}

\journal{Neurocomputing}

\begin{document}

\begin{frontmatter}


% \author{Name\corref{cor1}\fnref{label2}}

\title{MetaSTH-Sleep: Towards Effective Few-Shot Sleep Stage Classification \textcolor{black}{for Health Management} with Spatial-Temporal Hypergraph Enhanced Meta-Learning}

\author[1]{Jingyu Li}
%\ead{lijy1970@126.com}

\affiliation[1]{organization={The First Affiliated Hospital of Zhengzhou University},%Department and Organization
            city={Zhengzhou},
            country={China}}

\author[2]{Tiehua Zhang\corref{cor1}}
\ead{tiehuaz@tongji.edu.cn}
\cortext[cor1]{Corresponding authors:}
\affiliation[2]{organization={School of Computer Science and Technology, Tongji University},%Department and Organization
            city={Shanghai},
            country={China}}

\author[3]{Jinze Wang\corref{cor1}}
\ead{jinzewang@swin.edu.au}
\affiliation[3]{organization={School of Engineering, Swinburne University of Technology},%Department and Organization
            city={Melbourne},
            country={Australia}}

\author[4]{Yi Zhang}
\affiliation[4]{organization={Heart Center, Shanghai Tenth People’s Hospital, School of Medicine, Tongji University},%Department and Organization
            city={Shanghai},
            country={China}}

\author[5]{Yuhuan Li}
%\ead{li.yuhuan1@zs-hospital.sh.cn}
\affiliation[5]{organization={Liver Cancer Institute, Zhongshan Hospital, Key Laboratory of Carcinogenesis and Cancer Invasion, Fudan University},%Department and Organization
            city={Shanghai},
            country={China}} 

\author[4]{Yifan Zhao}
%\ead{zhaoyifan47@163.com}

\author[6]{Zhishu Shen}
%\ead{z\_shen@ieee.org}
\affiliation[6]{organization={School of Computer Science and Artificial Intelligence, Wuhan University of Technology},%Department and Organization
            city={Wuhan},
            country={China}}

\author[7]{Libing Wu}
%\ead{z\_shen@ieee.org}
\affiliation[7]{organization={School of Cyber Science and Engineering, Wuhan University},%Department and Organization
            city={Wuhan},
            country={China}}

\author[8,9,10]{Jiannan Liu}
%\ead{laurence\_ljn@163.com}
\affiliation[8]{organization={Department of Oral and Maxillofacial Head and Neck Oncology, Shanghai Ninth People’s Hospital, Shanghai Jiao Tong University School of Medicine},%Department and Organization
            city={Shanghai},
            country={China}} 
\affiliation[9]{organization={College of Stomatology, Shanghai Jiao Tong University},%Department and Organization
            city={Shanghai},
            country={China}} 
\affiliation[10]{organization={National Center for Stomatology},%Department and Organization
            city={Shanghai},
            country={China}}

%% Abstract
\begin{abstract}
\textcolor{black}{Accurate classification of sleep stages based on bio-signals is fundamental not only for automatic sleep stage annotation, but also for clinical health management and continuous sleep monitoring.} Traditionally, this task relies on experienced clinicians to manually annotate data, a process that is both time-consuming and labor-intensive. In recent years, deep learning methods have shown promise in automating this task. However, three major challenges remain: (1) deep learning models typically require large-scale labeled datasets, making them less effective in real-world settings where annotated data is limited; (2) significant inter-individual variability in bio-signals often results in inconsistent model performance when applied to new subjects, limiting generalization; and (3) existing approaches often overlook the high-order relationships among bio-signals, failing to simultaneously capture signal heterogeneity and spatial-temporal dependencies. To address these issues, we propose MetaSTH-Sleep, a few-shot sleep stage classification framework based on spatial-temporal hypergraph enhanced meta-learning. Our approach enables rapid adaptation to new subjects using only a few labeled samples, while the hypergraph structure effectively models complex spatial interconnections and temporal dynamics simultaneously in EEG signals. Experimental results demonstrate that MetaSTH-Sleep achieves substantial performance improvements across diverse subjects, offering valuable insights to support clinicians in sleep stage annotation.
\end{abstract}

% %\includegraphics{Figures/overall.pdf}

\begin{highlights}
\item Designs a meta spatial-temporal hypergraph construction to capture high-order relations.  
\item Multi-head attention adaptively integrates spatial and temporal hyperedge information.  
\item Enables fast subject adaptation with few labeled samples. 
\item Experiments on ISRUC and UCD datasets show superior accuracy and generalization.  
\end{highlights}

%% Keywords

\end{frontmatter}



\section{Introduction}\label{sec: Intro}
Accurate classification of sleep stages plays a vital role in diagnosing sleep disorders in clinical settings, improving sleep quality, and supporting long-term health management. In recent years, medical practitioners have primarily focused on developing Polysomnography (PSG) based techniques to capture physiological signals from various human organs, aiming to accurately delineate different sleep stages. Key signals recorded by PSG include electroencephalography (EEG), electrooculography (EOG), electromyography (EMG), and electrocardiography (ECG). Due to the complex interactions between these organs and the influence of successive timestamps, the observations are interconnected, making the PSG data inherently spatial-temporal~\cite{boostani2017comparative,liu2024exploiting}. This classification involves analyzing time series data alongside spatial considerations. Apart from that, PSG captures diverse spectrograms from multiple organs, highlighting the necessity of leveraging multimodal signals from various sources, such as EOG and ECG. Nonetheless, the classification of sleep stages using multimodal signals poses significant challenges, primarily due to the intricate nature of inter-organ interactivity and the heterogeneous characteristics of the signal modalities. In a nutshell, interactivity refers to the continuous interplay between human organs during sleep~\cite{ivanov2021signal}, while heterogeneity encompasses the variations in spectrograms observed across different signal types~\cite{zhang2023adaptive}.

With the advancement of deep learning techniques, researchers have found that these methods are more effective in decoding the intricate correlations between signal dynamics and sleep stages. Recently, various deep learning-based approaches~\cite{fiorillo2019automated}, such as Convolutional Neural Networks (CNNs) and Recurrent Neural Networks (RNNs), have been extensively investigated to address the sleep stage classification problem. The core principle of these works lies in extracting the feature representation of the input data from the time
domain, frequency domain, and time-frequency domain, which is then processed through a multilayer perceptron to map the learned features to corresponding sleep stages. This strategy significantly improves classification efficiency and accuracy, offering notable advantages over conventional, manually-driven approaches and great clinic values. Despite promising performance in sleep stage classification, CNNs and RNNs are constrained by their reliance on grid-like input formats, which fail to capture the intrinsic non-Euclidean structure of brain connectivity. 

Since brain regions are functionally interconnected in a non-Euclidean space, graph-based representations are considered more appropriate for capturing these spatial relationships. Inspired by the success of the graph convolution network (GCN) model in graph data, the work in~\cite{jia2020graphsleepnet} first used GCN to learn the spatial-temporal feature of EEG waves, where each EEG channel corresponds to a node of the graph at each timestamp, and the connection between the channels correspond to the edge of that graph. Building on this foundation, recent studies have concentrated on the design of spatial-temporal graph learning frameworks that jointly model the spatial interconnections and temporal dynamics inherent in EEG waves~\cite{feng2022eeg}. Empirical results indicate that these methods outperform traditional deep learning architectures such as CNNs and RNNs in sleep stage classification tasks. While spatial-temporal graph learning is proven effective to model both spatial dependencies among EEG channels (reflecting brain region connectivity) and temporal dynamics of EEG signals, researchers have also noted that constructing an effective and biologically meaningful graph remains a non-trivial task. Specifically, most spatial-temporal graph learning methods in prior research either utilize static measures, such as the phase-locking value (PLV), to derive a fixed graph structure from EEG data, or they employ trainable coefficients to determine the connectivity among nodes. Nevertheless, these works put efforts in exploiting the pairwise connectivity in both the spatial and temporal dimensions within the graph structure, which essentially oversimplify the brain's actual network behavior and fail to model higher-order relationships (e.g., interactions among three or more brain regions simultaneously).

Moreover, deep learning methods generally require large volumes of data to attain high classification performance. However, in practical clinical settings, acquiring sufficient data is often challenging due to resource constraints or variability across datasets, leading to the issue of data scarcity~\cite{thapa2024sleepfm}. Furthermore, models trained on data from a specific patient (subject) often fail to generalize to new individuals due to data imbalance caused by variations in factors such as the number and placement of EEG channels, sampling rates, experimental protocols, and subject demographics~\cite{eldele2023self}. These challenge greatly limits the direct applicability and scalability of conventional deep learning techniques in clinical settings. To address these limitations, few-shot learning (FSL) has emerged as a promising paradigm, enabling effective model training with only a small number of samples~\cite{banluesombatkul2020metasleeplearner}. Among various FSL approaches, meta-learning has attracted particular attention for its ability to enhance adaptability by focusing on learning how to learn, rather than merely fitting specific task~\cite{wang2024leveraging}. Meta-learning improves traditional deep learning by optimizing the model initialization, allowing for rapid adaptation to new tasks with minimal data. Typically, the training process involves constructing tasks composed of an $N$-way $K$-shot support set for feature extraction and a corresponding query set for evaluation, where $N$ denotes the number of classes and $K$ represents the number of samples per class. By maintaining relatively small $N$ and $K$ values, meta-learning methods are especially well-suited for scenarios with limited data and enable rapid adaptation to new subjects with varying individual characteristics, thereby significantly enhancing model generalization in clinical applications \textcolor{black}{and broadening its utility for health management.}

\begin{figure}[t]
    \centering
    \begin{subfigure}{0.5\textwidth}
    \includegraphics[width=\linewidth]{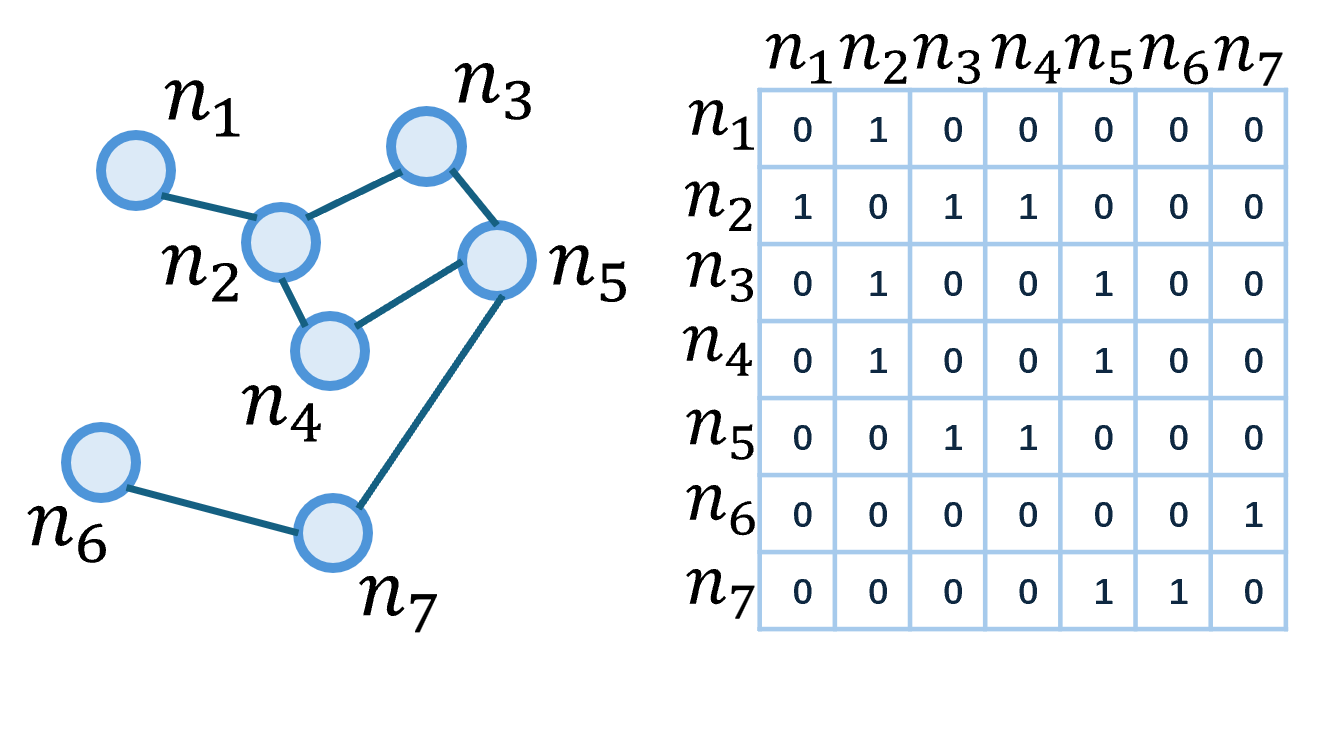}
    \caption{}
    \label{fig:graph_chart}
    \end{subfigure}
    \begin{subfigure}{0.4\textwidth}
    \includegraphics[width=\linewidth]{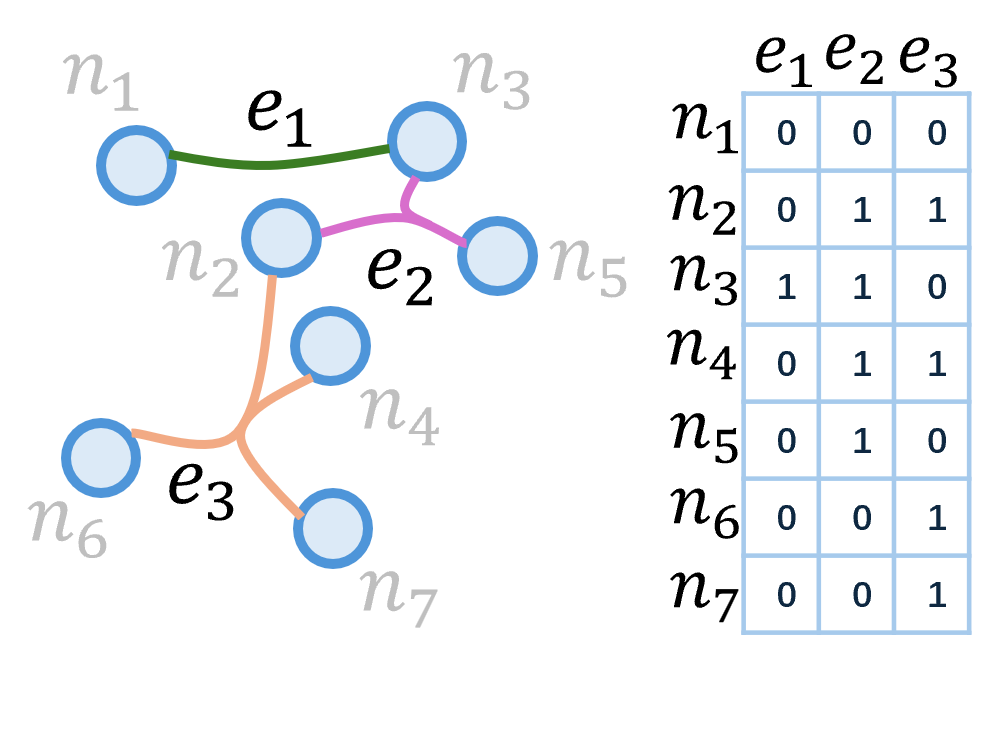}
    \caption{}
    \label{fig:hyper_chart}
    \end{subfigure}   
    \caption{Illustration of the structural difference between traditional graphs and hypergraphs. (a) A graph with respective adjacency matrix, which models only pairwise relationships among nodes. (b) A hypergraph and its incidence
matrix, where each hyperedge can connect multiple nodes simultaneously, enabling the modeling of high-order relationships.}
    \label{Fig:Graph_HyperGraph}
\end{figure}

%\improve{(this paragraph should describe why we need few shot and meta learning in here, the motivation)}.
Compared with traditional graph learning methods, where each edge connects only two nodes, hypergraphs introduce hyperedges that can simultaneously connect multiple nodes, enabling the capture of higher-order relationships. Fig.~\ref{Fig:Graph_HyperGraph} illustrates the structural difference between traditional graphs and hypergraphs. Fig.~\ref{fig:graph_chart} shows a graph and its corresponding adjacency matrix, which captures only pairwise relationships among nodes. Fig.~\ref{fig:hyper_chart} shows a hypergraph and its incidence matrix, where each hyperedge can connect multiple nodes simultaneously, enabling the modeling of high-order dependencies. In the context of sleep stage classification, such high-order relationships often emerge from joint interactions across multiple physiological channels (e.g., EEG, EOG, EMG). Hypergraphs thus provide a more expressive framework for representing the intrinsic spatial-temporal structures in complex biomedical data.
%\shen{[WHAT IS HYPERGRAH?]
To this end, we propose MetaSTH-Sleep, a few-shot sleep stage classification framework based on spatial-temporal hypergraph enhanced meta-learning. This framework consists of two key modules: a dynamic spatial-temporal hypergraph designed to extract higher-order relationships, capturing both spatial interdependencies and temporal dynamics, and a meta-learning module that facilitates rapid and efficient adaptation to new tasks with limited data. The contributions of this paper are summarized as follows:

\begin{itemize}
   \item We propose a novel spatial-temporal hypergraph enhanced meta-learning framework, namely MetaSTH-Sleep, for few-shot sleep stage classification. The framework facilitates rapid and efficient adaptation to new subjects under limited data condition, addressing the challenge of data scarcity.
   \item To the best of our knowledge, this is the first work to integrate spatial-temporal hypergraph into meta-learning paradigm for sleep stage classification. The framework is capable of capturing the hidden spatial correlations and temporal dependencies from the EEG waves simultaneously through dynamic hyperedge construction and attention-based embedding updates.
   \item Extensive experiments on two real-world datasets validate the effectiveness of our framework, showing that MetaSTH-Sleep consistently outperforms existing state-of-the-art methods in sleep stage classification tasks.
\end{itemize}

\section{Related Work}\label{sec: relatedWork}
\subsection{Deep Learning for Spatial-Temporal EEG Signals}
The analysis of spatial-temporal EEG signals data has garnered considerable attention in recent years, fueled largely by the rapid progress in advanced deep learning methodologies. Previous studies have highlighted that the automated and efficient analysis of such data offers significant clinical value for medical practitioners~\cite{9530406}. Nevertheless, uncovering and interpreting the latent characteristics embedded within these numerical yet complex datasets remains a persistent and formidable challenge.

To address this, recent studies have focused on modeling spatial and temporal dependencies, enabling deep learning models to enhance prediction and classification performance. For example, the Fast Discriminative Complex-valued Convolutional Neural Network (FDCCNN)~\cite{zhang2017new} has been introduced to capture intricate correlations in EEG signals, significantly improving the accuracy of model-based sleep stage classification tasks. Similarly, the MLP-Mixer architecture~\cite{huang2022mlp} has demonstrated promising performance in handling multi-channel temporal signals for regression tasks. RNNs have also been widely adopted to model the non-linear interdependencies inherent in spatial-temporal data. Models such as ConvLSTM~\cite{shi2015convolutional} and Bi-LSTM~\cite{zheng2020hybrid} have proven effective in this regard. Building upon these advances, attention mechanisms have been incorporated to better capture long-term dependencies~\cite{ienco2020deep}, with spatial and temporal attention modules specifically crafted to cluster multivariate time series data of varying lengths more effectively.

Moreover, earlier studies have observed that time-invariant features exist within spatial dimensions~\cite{sun2019hierarchical,supratak2017deepsleepnet}, suggesting that distinct modeling strategies for spatial and temporal features could be beneficial. For instance, DeepSleepNet~\cite{supratak2017deepsleepnet} employs a two-step approach that leverages CNNs to extract spatial features, followed by bidirectional long short-term memory (Bi-LSTM) networks to model temporal transitions. Similarly, a hierarchical neural network proposed in~\cite{sun2019hierarchical} separates the learning of spatial representations and temporal sequences into distinct stages.
%using of GNN on EGG data
\subsection{Graph Neural Networks (GNNs) on EEG Data}
%\subsection{Recent Advances of Using Graph Neural Networks (GNNs)}%\shen{[THE TITLE SHOULD BE MODIFIED]}
While the application of CNNs and RNNs to spatial-temporal EEG data has achieved encouraging outcomes, certain limitations remain evident. Primarily, these approaches necessitate data processing within Euclidean space, thereby overlooking potential connectivity information inherent in the spatial dimension. Additionally, the interdependent relationships identified by these models can be challenging for domain experts to comprehend. The emergence of graph neural networks (GNNs) has introduced new possibilities in this area, prompting researchers to explore how graphs can better represent topological structures in both spatial and temporal dimensions~\cite{wu2020comprehensive}.
Despite the promising results achieved by GNNs on non-Euclidean data, constructing graph structures from EEG waves remains a non-trivial task. This transformation involves two primary challenges: first, identifying spatial correlations among different EEG electrode (channels) to create an adjacency matrix; second, extracting node features from temporal values. Several studies have endeavored to address these challenges~\cite{jain2016structural, yu2017spatio, seo2018structured, yan2018spatial}. For instance, the graph convolutional recurrent network (GCRN)~\cite{seo2018structured} integrates long short-term memory (LSTM) networks with Chebyshev networks (ChebNet) to process spatial-temporal data. Structural-RNN~\cite{jain2016structural} employs RNNs at both the node and edge levels to reveal spatial correlations within data. An alternative approach involves using CNNs to embed temporal relationships, mitigating issues like exploding or vanishing gradients. For example, the ST-GCN model~\cite{yan2018spatial} uses partitioned graph convolution layers to extract spatial features and one-dimensional convolution layers to capture temporal dependencies. Similarly, CGCN~\cite{yu2017spatio} combines a one-dimensional convolution layer with ChebNet or GCN layers to handle spatial-temporal data. However, previous research either utilizes static measures, such as the phase-locking value (PLV), to derive
a fixed graph structure from EEG data, or they employ trainable coefficients to generate the pairwise connectivity of nodes, which oversimplifies the higher-order relations existed in both spatial and temporal dimension, failing to encode the brain’s actual network behavior.

%mention in introduction
\subsection{Meta-learning for Data Scarcity and Imbalance} %\shen{[IS IT MENTION WELL IN SECTION 1?]}}
With the emergence of GNNs, significant progress has been made in modeling spatial-temporal EEG data. However, several limitations persist. Notably, most GNN-based approaches rely heavily on large amounts of data. In practical clinical applications, acquiring such extensive datasets is often infeasible due to factors such as limited resources and patient privacy concerns, leading to the issue of data scarcity~\cite{thapa2024sleepfm}. Additionally, models trained on data from specific subjects often struggle to generalize to unseen individuals, primarily due to data imbalance caused by inter-subject variability~\cite{eldele2023self}. These challenges hinder the scalability and general applicability of GNN-based methods in clinical settings.

In this context, model-agnostic meta-learning (MAML) has attracted increasing attention for its ability to learn generalizable knowledge across tasks, enabling effective knowledge transfer from only a small amount of data~\cite{finn2017model,wang2023meta}. For example, FL-ML~\cite{moon2023federated} has been investigated to validate that initializing models with pre-trained weights from the MAML model yields faster and more accurate performance in sleep stage classification compared to random initialization. Similarly, MetaSleepLearner (MSL)~\cite{banluesombatkul2020metasleeplearner} was introduced to transfer sleep stage knowledge acquired from a large dataset to new individuals in unseen cohorts, using a MAML framework combined with CNNs. S2MAML~\cite{lemkhenter2022towards} was proposed to generalize across different patients and recording facilities by building on the MAML framework and incorporating a self-supervised learning (SSL) stage. DA-RelationNet~\cite{an2023dual} was introduced to learn representative features of unseen subject categories and classify them using limited EEG data by integrating a MAML based temporal-attention module. Compared with these methods, we propose to explicitly integrate spatial-temporal correlations into the MAML framework, enabling the model to simultaneously capture both spatial dependencies and dynamic temporal patterns inherent in EEG signals. By doing so, our method is better equipped to generalize across subjects with high inter-individual variability, especially under the constraints of data scarcity common in clinical settings.

\begin{figure}[t]
    \centering
    \includegraphics[width=\textwidth]{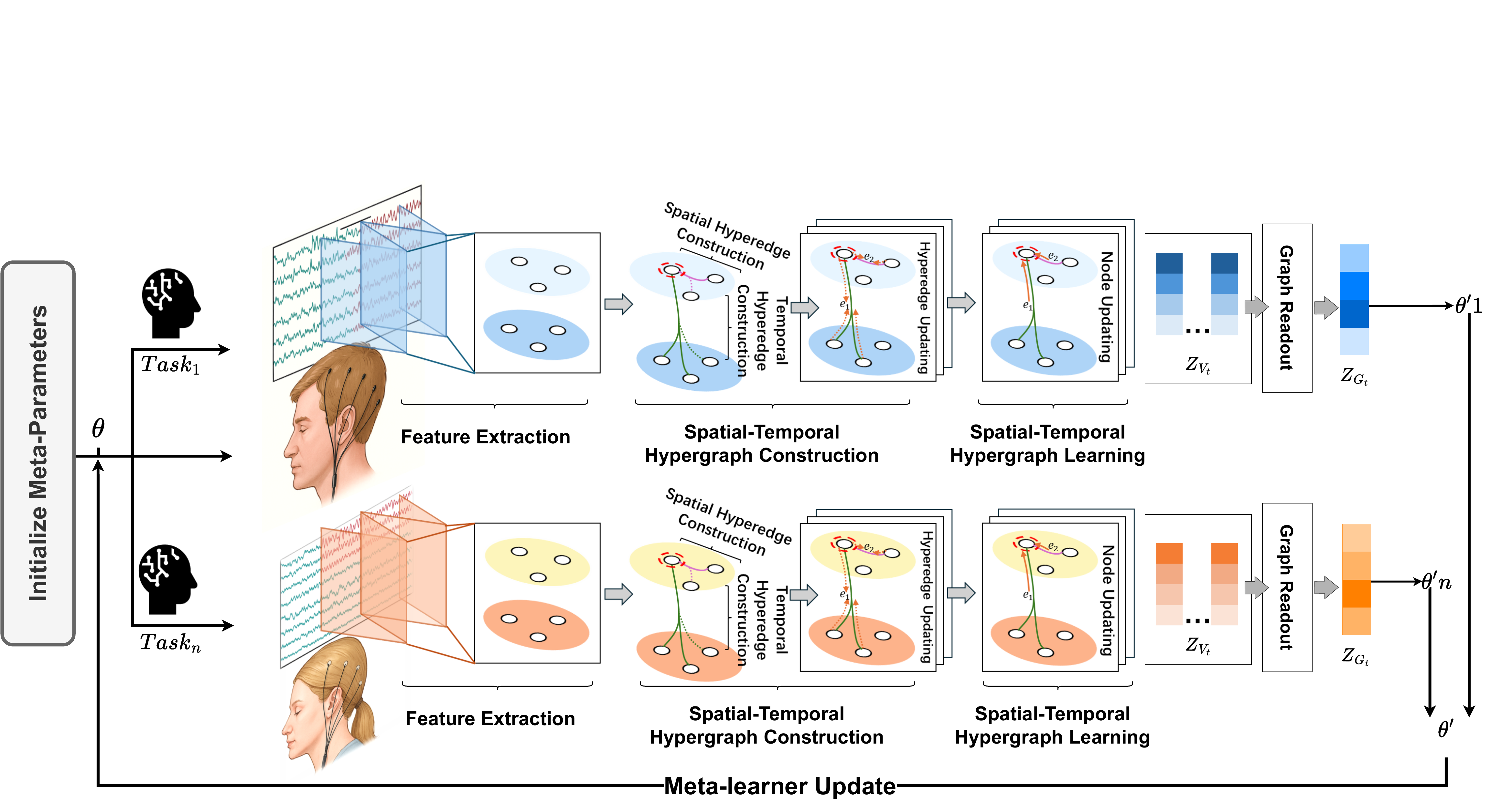}
    \caption{An overview of MetaSTH-Sleep.}
    \label{fig:overall}
\end{figure}

\section{Methodology}\label{sec: method}
%\improve{better add a psudo algorithm, now this sec is too short}

This section introduces a few-shot learning framework based on spatial-temporal hypergraph enhanced meta-learning (MetaSTH-Sleep) for sleep stage classification. The proposed method aims to address the challenge of individual variability in physiological signals by enabling fast adaptation of a sleep stage classifier to new subjects with only a few labeled samples. The overall architecture of MetaSTH-Sleep is shown in Fig.~\ref{fig:overall}. %\shen{[MORE DETAILS ON OVERVIEW INFO ARE REQUIRED + OVERVIEW FIG]}

Let $\mathbf{X} \in \mathbb{R}^{T \times N \times d}$ denote the input sequence of multimodal physiological signals, where $T$ is the number of time steps, $N$ is the number of channels (e.g., EEG), and $d$ is the feature dimension per channel at each time step. At each time $t$, we define the input sample as $\mathbf{X}_t = \{\mathbf{x}_t^1, \dots, \mathbf{x}_t^N\} \in \mathbb{R}^{N \times d}$, where $\mathbf{x}_t^i \in \mathbb{R}^d$ is the feature of the $i$-th channel at time $t$. To model temporal context, we consider a pair of adjacent time steps $(t-1, t)$, denoted as $\mathbf{X}_{t-1:t} = \{\mathbf{X}_{t-1}, \mathbf{X}_t\} \in \mathbb{R}^{2 \times N \times d}$. Each time step $t$ has an associated ground truth label $y_t \in \{1, \dots, C\}$, where $C$ is the number of sleep stages.

\subsection{Model-Agnostic Meta-Learning (MAML)}

We adopt MAML to learn a meta-parameter $\theta_0$ that enables rapid adaptation to new subjects. Each task $\mathcal{T}_b$ corresponds to one subject’s PSG recording, split into a support set $\mathcal{D}_b^{\text{spt}}$ and a query set $\mathcal{D}_b^{\text{qry}}$.

Given the base learner $f_\theta$ with parameters $\theta$, MAML performs inner-loop adaptation on $\mathcal{D}_b^{\text{spt}}$:
\begin{equation}
\theta_b' = \theta_0 - \alpha \nabla_\theta \mathcal{L}_{\mathcal{T}_b}^{\text{spt}}(f_\theta),
\end{equation}
where $\alpha$ is the inner-loop learning rate and $\mathcal{L}_{\mathcal{T}_b}^{\text{spt}}$ is the task-specific loss. The outer-loop minimizes the query loss after adaptation:
\begin{equation}
\mathcal{L}_{\text{meta}}(\theta_0) = \sum_{b=1}^{B} \mathcal{L}_{\mathcal{T}_b}^{\text{qry}}(f_{\theta_b'}),
\label{Eq：loss}
\end{equation}
where $B$ is the number of sampled tasks per meta-batch. The meta-parameters are then updated by:
\begin{equation}
\label{eq:meta}
\theta_0 \leftarrow \theta_0 - \beta \nabla_{\theta_0} \mathcal{L}_{\text{meta}}(\theta_0),
\end{equation}
where $\beta$ is the meta learning rate.

\begin{figure}[tbp]
    \centering
    \includegraphics[width=0.7\textwidth]{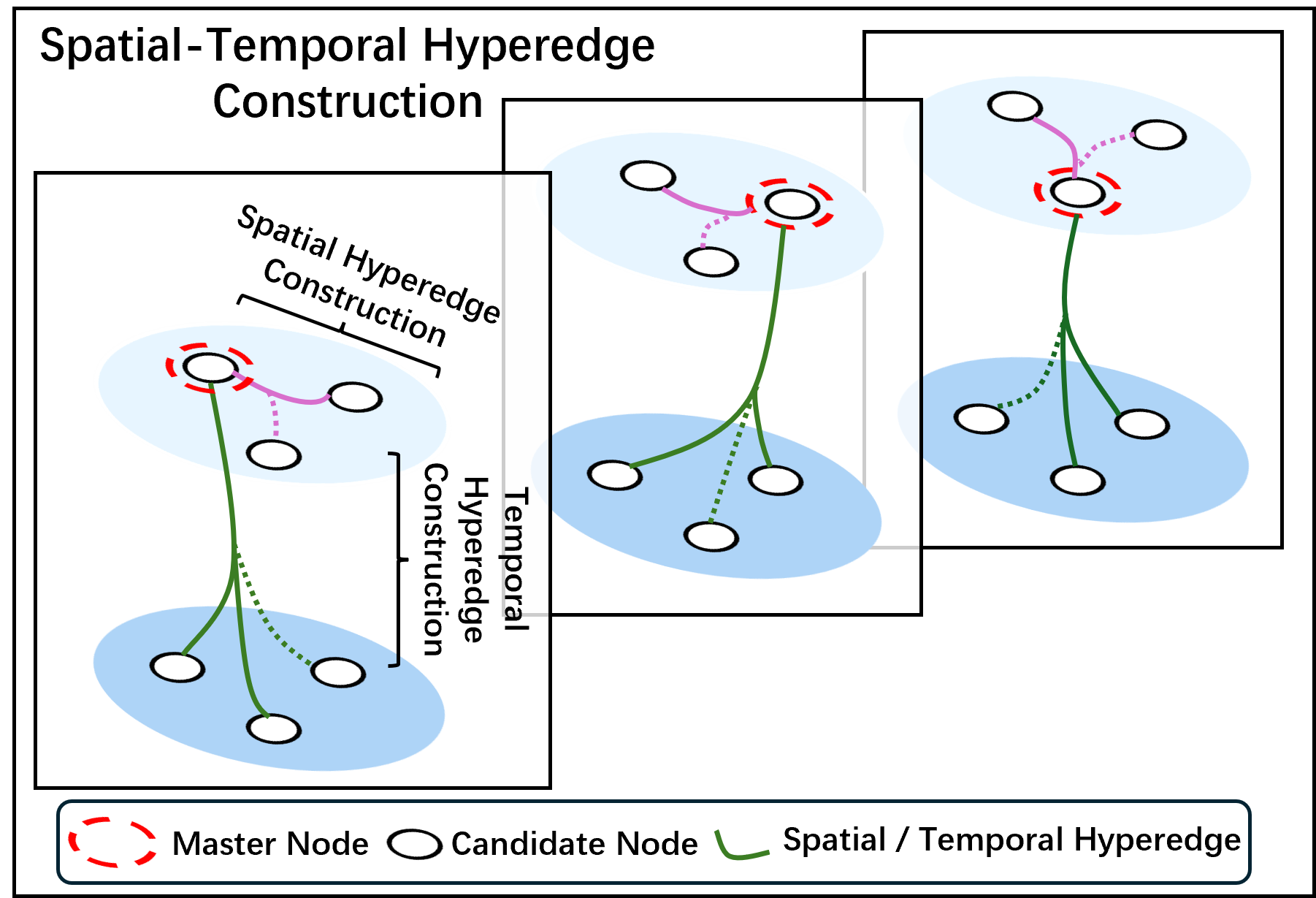}
    \caption{Illustration of the construction process of spatial and temporal hyperedges.}
\label{fig:hypergraph_construction}
\end{figure}

\subsection{Meta Spatial-Temporal Hypergraph Construction}

To capture spatial and temporal correlations, we construct a spatial-temporal hypergraph at each time step $t$ as shown in Fig.~\ref{fig:hypergraph_construction}. Each node corresponds to a channel-specific feature vector from either $t$ or $t-1$. The node set is defined as $\mathcal{V}_t = \{v^{t-1}_1, \dots, v^{t-1}_N, v^{t}_1, \dots, v^{t}_N\}$. The hypergraph is denoted as $\mathcal{G}_t = (\mathcal{V}_t, \mathcal{E}_t)$, which includes two types of hyperedges: spatial hyperedges $\mathcal{E}_t^{\text{spa}}$ connect nodes within the same time step, while temporal hyperedges $\mathcal{E}_t^{\text{tem}}$ connect nodes across $t-1$ and $t$.

Hyperedges are constructed dynamically via a reconstruction-based mechanism. For a given master node $\dot{v}$, the spatial reconstruction error is defined as:
\begin{equation}
c_{\text{spa}}(\dot{v}) = \left\| \mathbf{x}_{\dot{v}} \Theta_{\text{spa}} - \mathbf{p}_{\dot{v}}^{\text{spa}} \mathbf{X}_{\text{spa}} \right\|_2^2,
\end{equation}
where $\Theta_{\text{spa}} \in \mathbb{R}^{d \times d'}$ is a projection matrix, $\mathbf{p}_{\dot{v}}^{\text{spa}} \in \mathbb{R}^{|S_{\text{spa}}|}$ is a learnable reconstruction coefficient vector, and $\mathbf{X}_{\text{spa}}$ is the feature matrix of candidate nodes. Nodes with $p_i > 0$ are selected into the spatial hyperedge $e_{\text{spa}}(\dot{v})$. Temporal hyperedges $e_{\text{tem}}(\dot{v})$ are constructed analogously.

This reconstruction-based mechanism allows each hyperedge to simultaneously connect a master node with multiple candidate nodes, thereby encoding higher-order relationships. In contrast to conventional graph construction, which is restricted to pairwise connectivity, the hyperedge formulation captures the joint contribution of several nodes to the representation of the master node. Such a mechanism provides a more expressive description of complex spatial-temporal co-activation patterns across EEG channels.

The total reconstruction loss is:
\begin{equation}
\mathcal{L}_{\text{recon}} = \sum_{\dot{v} \in \mathcal{V}_t} \lambda \left(c_{\text{spa}}(\dot{v}) + c_{\text{tem}}(\dot{v})\right)
+ \left\| \mathbf{p}_{\dot{v}}^{\text{spa}} \right\|_1 + \left\| \mathbf{p}_{\dot{v}}^{\text{tem}} \right\|_1
+ \gamma \left( \left\| \mathbf{p}_{\dot{v}}^{\text{spa}} \right\|_2^2 + \left\| \mathbf{p}_{\dot{v}}^{\text{tem}} \right\|_2^2 \right).
\end{equation}

\subsubsection{Hyperedge Embedding Update}
\begin{figure}
    \centering
    \includegraphics[width=0.7\textwidth]{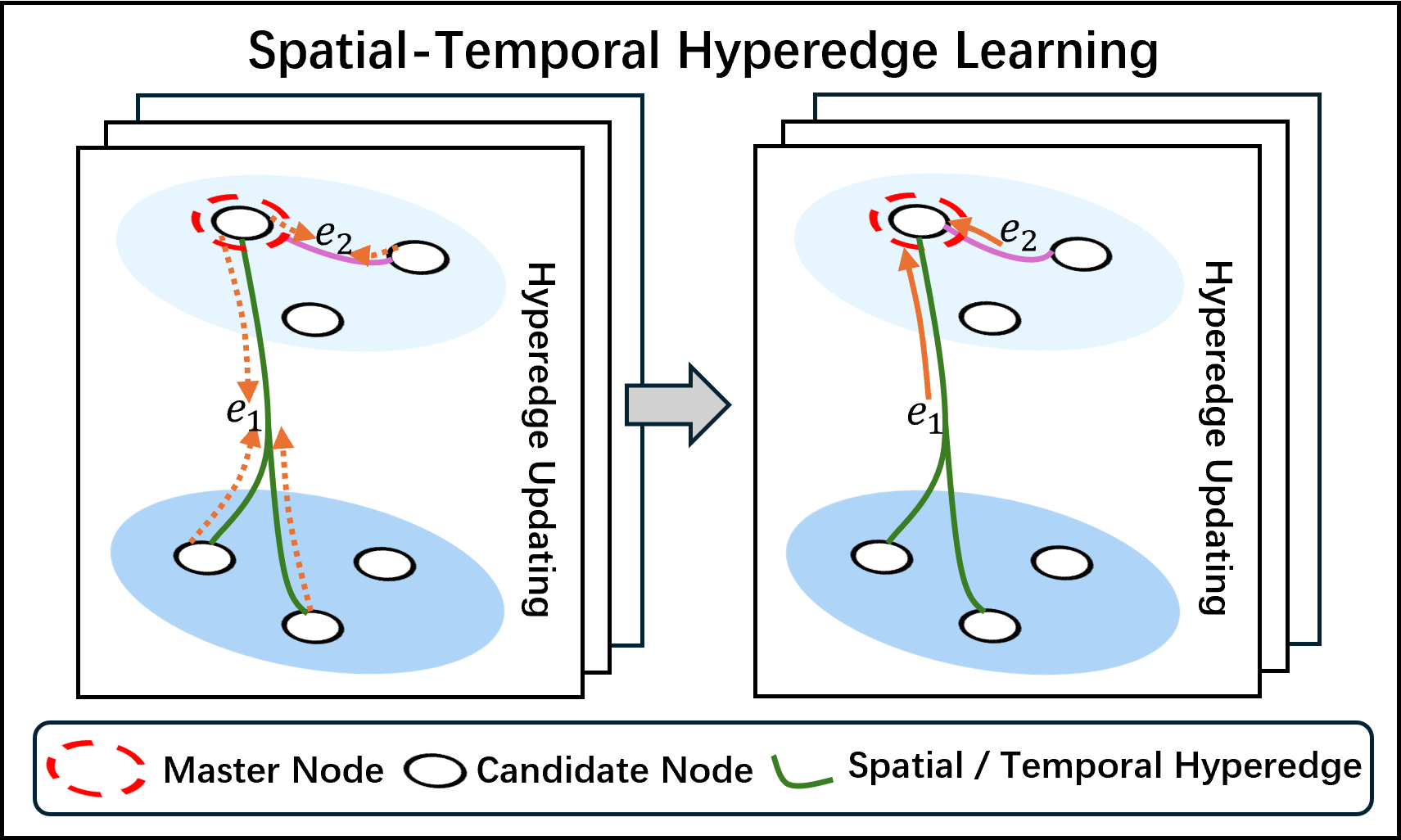}
    \caption{Illustration of the process of spatial-temporal hyperedge learning.}
    \label{fig:sthlearning}
\end{figure}

The process of spatial and temporal learning is shown in Fig.~\ref{fig:sthlearning}. Let $H \in \mathbb{R}^{|\mathcal{V}_t| \times |\mathcal{E}_t|}$ be the incidence matrix of the hypergraph. The embedding of each hyperedge $e \in \mathcal{E}_t$ is:
\begin{equation}
E(e) = \frac{\sum_{v \in \mathcal{V}_t} H(v,e) \cdot \mathbf{x}_v}{\sum_{v \in \mathcal{V}_t} H(v,e)},
\end{equation}
where $H(v,e) = 1$ for the master node, and $H(v,e) = p_v$ for candidate nodes.

\subsubsection{Multi-head Node Embedding Update}
Each node $\dot{v} \in \mathcal{V}_t$ is associated with one spatial and one temporal hyperedge. For each attention head $h$, the attention score is:
\begin{equation}
\text{att}^h(\dot{v}, e) = \frac{(\mathbf{x}_{\dot{v}} Q^h) \cdot (E(e) K^h)^\top}{\sqrt{d_a}},
\end{equation}
where $Q^h, K^h \in \mathbb{R}^{d \times d_a}$ are projection matrices and $d_a$ is the attention dimension.

The final node embedding is:
\begin{equation}
z_{\dot{v}} = \text{MLP}\left( \big\|_{h=1}^{K} \left( w^{h}_{\text{spa}} \cdot V^{h}_{\text{spa}} + w^{h}_{\text{tem}} \cdot V^{h}_{\text{tem}} \right) \right),
\end{equation}
where $w^h$ are normalized attention weights, and $V^h$ are value embeddings. The graph-level embedding is:
\begin{equation}
Z_{\mathcal{G}_t} = \frac{1}{|\mathcal{V}_t|} \sum_{v \in \mathcal{V}_t} z_v.
\end{equation}
%combine 3.2/3.3

\begin{algorithm}[ht]
\caption{Meta-training of MetaSTH-Sleep}
\label{alg:metasth}
\small
\begin{algorithmic}[1] % 1 means line numbers
\State \textbf{Input:} Meta tasks $\{\mathcal{T}_b\}_{b=1}^{B}$; meta-parameters $\theta_0$; learning rates $\alpha, \beta$; time $t$. \State \textbf{Output:} Optimized meta-parameters $\theta_0$
\State Initialize $\theta_0$ randomly
\For{each meta-training iteration}
    \State Sample a batch of tasks $\{\mathcal{T}_b\}_{b=1}^{B}$
    \For{each task $\mathcal{T}_b$}
        \State Split $\mathcal{T}_b$ into support set $\mathcal{D}_b^{\text{spt}}$ and query set $\mathcal{D}_b^{\text{qry}}$
        \For{each sample $\in \mathcal{D}_b^{\text{spt}}$ }
            \State Build spatial-temporal hypergraph $\mathcal{G}_t = (\mathcal{V}_t, \mathcal{E}_t)$
            \State Construct hyperedges via reconstruction
            \State Compute hyperedge embeddings
            \State Update node embeddings $z_{\dot{v}}$ via multi-head attention
            \State Aggregate nodes $\in \mathcal{V}_t$ to get $Z_{\mathcal{G}_t}$
            \State Predict sleep stage $\hat{y}_t$ from $Z_{\mathcal{G}_t}$
        \EndFor
        \State Compute support loss $\mathcal{L}_{\mathcal{T}_b}^{\text{spt}} = \alpha \mathcal{L}_{\text{recon}} + (1-\alpha) \mathrm{CE}(\hat{y}_t, y_t)$
        \State Update adapted parameters $\theta_b' = \theta_0 - \alpha \nabla_\theta \mathcal{L}_{\mathcal{T}_b}^{\text{spt}}$
    \EndFor

    \For{each task $\mathcal{T}_b$}
        \State Compute query loss $\mathcal{L}_{\mathcal{T}_b}^{\text{qry}}$ using Eq.~\ref{Eq：loss}
    \EndFor

    \State Aggregate meta-objective: $\mathcal{L}_{\text{meta}} = \sum_{b=1}^{B} \mathcal{L}_{\mathcal{T}_b}^{\text{qry}}$
    \State Update meta-parameters: $\theta_0 \leftarrow \theta_0 - \beta \nabla_{\theta_0} \mathcal{L}_{\text{meta}}$
\EndFor
\end{algorithmic}
\end{algorithm}
\subsubsection{Hypergraph-based Learner}

The hypergraph-based learner $f_\theta$ is optimized using the MAML framework. The per-task loss in the inner loop is:
\begin{equation}
\mathcal{L}_{\mathcal{T}_b}(f_\theta) = \alpha \mathcal{L}_{\text{recon}} + (1 - \alpha) \cdot \mathrm{CE}(\hat{y}_t, y_t),
\end{equation}
where $\mathrm{CE}$ denotes cross-entropy and $\alpha \in [0,1]$ balances the reconstruction and classification terms.

Accordingly, the meta-update in Eq.~\ref{eq:meta} becomes:
\begin{equation}
\theta_0 \leftarrow \theta_0 - \beta \nabla_{\theta_0} \left( \alpha \mathcal{L}_{\text{recon}} + (1 - \alpha) \cdot \mathrm{CE}(\hat{y}_t, y_t) \right).
\end{equation}
Algorithm.~\ref{alg:metasth} shows the meta-training process of MetaSTH-Sleep.

\subsection{Complexity Analysis}
In the process of spatial-temporal hypergraph generation, the complexity of algorithm mainly lies in the reconstruction of each master node from its candidate node set. As hyperedges of both spatial and temporal types are generated for every node, the cost of hyperedge generation is $O (N^2 d' + N d d'),$
where $N$ is the number of nodes, $d$ is the input feature dimension, and $d'$ is the projected dimension. Additionally, during hyperedge embedding update (Eq.~(6)), the incidence matrix $H$ and node embeddings $X$ are used to compute hyperedge embeddings, with computational cost $O (N^2 d)$. Moreover, the multi-head attention node update (Eq.~(7)–(9)) requires computing queries and keys for each node–hyperedge pair across $K$ heads with attention dimension $d_a$, leading to complexity $O (K N d d_a).$ Hence, the total computational cost per time step is $O\!\big (N^2(d + d') + K N d d_a\big).$ Within the MAML framework, each task $\mathcal{T}_b$ undergoes inner-loop adaptation on its support set and outer-loop evaluation on its query set. Each forward and backward pass through the hypergraph learner maintains $O (N^2(d + d'))$ complexity. If $B$ tasks are sampled per meta-batch and the number of adaptation steps is fixed, the total computational cost per meta-training iteration scales as $O\!\Big (B \,[|\mathcal{D}_b^{\text{spt}}| + |\mathcal{D}_b^{\text{qry}}|]\; N^2(d + d')\Big).$ Since $B$ and the number of adaptation steps are typically small constants in practice, the overall complexity remains on par with many existing hypergraph-based models, such as~\cite{liu2024exploiting} and~\cite{lou2021stfl} .

\section{Experiments}\label{sec: Experiments}
We conduct extensive experiments on real-word datasets to evaluate our proposed method from different perspectives. We aim to answer the following research questions: %\shen{[IT IS BETTER TO MENTION RQ EARLIER, E.G., SECTION 1 TO CLARIFY THE MOTIVATION]}:
% Due to the difficulty in acquiring high-quality data in real-world medical environments, the problem of limited data availability is common. Moreover, models trained on one subject often fail to generalize directly to another because of variations arising from multiple factors, such as the number and placement of EEG channels, sampling frequencies, experimental protocols, and subject types. These challenges significantly hinder the performance of sleep stage classification. In this section, we conduct a series of experiments on real-world datasets to evaluate the effectiveness of the proposed method under few-shot learning scenarios.
\begin{itemize}
    \item \textbf{RQ1:} How does our proposed method perform in terms of overall classification accuracy and per-class performance compared to baselines?
    \item \textbf{RQ2:} How do different methods handle class-level confusion, and where do they succeed or fail?
    \item \textbf{RQ3:} How robust is our method under inter-subject variability?
    \item \textbf{RQ4:} How does the model's performance vary under sensitivity to adaptation steps and support set size in few-shot scenario?
    \item \textbf{RQ5:} How do different components, such as spatial-temporal hypergraph construction and multi-head attention, affect the performance of MetaSTH-Sleep?
\end{itemize}
%RQ4/5 

\subsection{Dataset}
In our experiments, we employ two widely recognized benchmark datasets, ISRUC~\cite{khalighi2016isruc} and UCD~\cite{goldberger2000physiobank}, to evaluate the performance of the proposed method. The ISRUC dataset, introduced in 2015 by the Sleep Medicine Centre at Coimbra University Hospital (CHUC), comprises a total of 118 polysomnography (PSG) recordings. It is divided into three subgroups: recordings from 100 subjects with a history of sleep disorders captured in a single session, recordings from 8 subjects across two separate sessions, and recordings from 10 healthy individuals. Following~\cite{liu2024exploiting}, we utilize Subgroup 3, which contains recordings from 10 healthy participants collected in a single session, making it particularly suitable for comparative studies between healthy subjects and those with sleep disorders. Each PSG recording is annotated according to the American Academy of Sleep Medicine (AASM) guidelines, covering five sleep stages: Wake, N1, N2, N3, and REM (Rapid Eye Movement). The UCD dataset, formally known as the St. Vincent’s University Hospital / University College Dublin Sleep Apnea Database, was revised in 2011. It contains overnight PSG recordings from 25 subjects (21 males and 4 females) diagnosed or suspected of sleep-related breathing disorders, including obstructive sleep apnea (OSA), central sleep apnea, and habitual snoring. Sleep stages were initially scored by sleep specialists according to the Rechtschaffen and Kales (R\&K) rules, resulting in eight categories: Wake, Stage 1, Stage 2, Stage 3, Stage 4, REM, Artifact, and Indeterminate. In alignment with the AASM standards, S3 and S4 were merged into a single stage termed slow-wave sleep (SWS), and segments labeled as Artifact and Indeterminate were excluded from this study following~\cite{banluesombatkul2020metasleeplearner}. Thus, the final considered stages in UCD dataset include Wake, S1, S2, SWS, and REM. The details of two datasets are summarized in Table~\ref{tab:datasets}.

\subsection{Experimental Settings}
To verify the model's ability to rapidly adapt to new tasks, we conduct experiments in which, for both datasets, each subject is sequentially selected as the test data, while the remaining subjects are used to construct the meta-training tasks. Following the few-shot learning paradigm~\cite{finn2017model, wang2025hyperman}, we adopt an $N$-way $K$-shot setting for the sleep stage classification task. Here, $N$ refers to the number of sleep stages, and $K$ denotes the number of labeled samples per class utilized for model adaptation. In our experiments, each subject is treated as a distinct meta-training task. For each task, we randomly sample $N \times K$ instances to form both the support set and the query set, resulting in a total of $2 \times N \times K$ samples per task during meta-training. For example, in the ISRUC dataset, where there are $N=5$ sleep stages, we randomly select $K=10$ samples per class for both the support and query sets, constructing a 5-way 10-shot problem. Thus, each task comprises $2 \times 5 \times 10 = 100$ samples. Since ISRUC contains data from 10 subjects, we allocate 9 subjects for meta-training and 1 subject for meta-testing, leading to $9 \times 100 = 900$ instances in total for meta-training. During the meta-testing phase, $K$ samples from each class are randomly selected from the unseen subject for fine-tuning. The primary objective of our experimental design is to assess the effectiveness of the fine-tuned model under the constraint of having only a limited number of labeled samples. To ensure a fair comparison, in subsequent experiments, we also constrain the data volume for non-meta-learning methods to match that of the few-shot setting. All reported results are obtained as the average over five independent runs with different random seeds and sampling (e.g., support set and query set).

\begin{table}[t]
\centering
\caption{Summary of the ISRUC and UCD datasets.}
\label{tab:datasets}
\footnotesize
\begin{tabular}{l|c|c}
\toprule
\hline
\textbf{Attribute} & \textbf{ISRUC (Subgroup 3)} & \textbf{UCD} \\
\hline
Source & CHUC Sleep Medicine Centre & St. Vincent's Hospital \\
Release Year & 2015 & 2011 \\
Number of Subjects & 10 (healthy only) & 25 (OSA and other disorders) \\
Sampling Type & PSG recordings & PSG recordings \\
Scoring Standard & AASM & R\&K (converted to AASM) \\
Annotation & Manual & Manual by specialists \\
Sleep Stages & Wake, N1, N2, N3, REM & Wake, S1, S2, SWS, REM \\
\hline
\bottomrule
\end{tabular}
\end{table}

\subsection{Comparison Methods} %\shen{[GIVE REFS TO THEM, AND I THINK MOST OF THE BASELINES ARE NOT NOVEL]}}
We compare the MetaSTH-Sleep with six state-of-the-art approaches. (1) Random Forest (RF)~\cite{huang2020sleep} is a classical ensemble learning method that constructs multiple decision trees and outputs the mode of their predictions. (2) Long Short-Term Memory (LSTM)~\cite{zhuang2022intelligent} is a recurrent neural network capable of modeling temporal dependencies in sequential data. (3) Convolutional Neural Network (CNN)~\cite{phan2018joint} is a deep learning model that captures local patterns via convolutional kernels. Used to extract spatial features from multichannel signals. (4) Graph Attention Network (GAT)~\cite{demir2022eeg} is a graph-based model that leverages attention mechanisms to learn node representations by focusing on important neighbors. (5) Model-Agnostic Meta-Learning (MAML)~\cite{finn2017model} is a meta-learning approach that learns initialization parameters allowing fast adaptation to new tasks with few examples. (6) MSL~\cite{banluesombatkul2020metasleeplearner} is a MAML-based few-shot learning model tailored for sleep stage classification, using CNNs as the base learner.

We adhere to the data preprocessing procedures and default model architectures as reported in the original works for all baseline methods described above. The optimal hyperparameters for each method are determined empirically based on their performance on a held-out validation set. Specifically, for baselines (1–4), the learning rate is searched over \{0.1, 0.05, 0.01, 0.005, 0.001, 0.0001\}. For meta-learning-based baselines (5–6), the inner-loop and outer-loop learning rates are searched over \{0.5, 0.1, 0.01, 0.001, 0.0001\}. For our proposed MetaSTH-Sleep model, the hyperparameters are configured as follows: the inner-loop learning rate $\alpha$ is set to 0.001, and the meta-learning rate $\beta$ is set to 0.0005. The number of sampled tasks per meta-batch $B$ is fixed at 3, and the number of adaptation steps is set to 3. A weight decay of 0.01 is applied for regularization, and a dropout rate of 0.3 is used during training.

\subsection{Experiment Results and Analysis}
\begin{table}[t]
\centering
\caption{Performance comparison on ISRUC and UCD datasets. \\(*The best results are highlighted in \textbf{bold}; the runner up is \underline{underlined}.)}
\label{Tab: 1}
\small
\begin{tabular}{c|c|ccccc}
\toprule
\hline
\multicolumn{7}{c}{ISRUC}                                             \\
\hline
                 & Accuracy & \multicolumn{5}{c}{F1 score per class} \\
                 &          & Wake & N1   & N2   & N3   & REM  \\
\hline
RF               & 0.5536        & 0.6050    & 0.1367    & 0.4934   & 0.8290    & 0.5335    \\
LSTM             & 0.5253        & 0.5342    &  0.3714    &  0.3802    & 0.7452    & 0.6663   \\
CNN              & 0.6183        & 0.6777    & 0.4663    & 0.6662    & 0.5464    & 0.7229    \\
GAT              & 0.5880        &  0.8303    & 0.4353    & 0.6276    & 0.0176    & 0.7275   \\
MAML             &  0.6355        & \underline{0.8423}    & 0.4006    & 0.5551    & 0.7181    & 0.5998    \\
MSL &  \underline{0.7747}        &  0.7901    & \underline{0.5494}    & \underline{0.7506}    & \underline{0.8560}    &  \underline{0.7612}    \\
MetaSTH-Sleep    & \textbf{0.8052}        & \textbf{0.8466}    &  \textbf{0.5803}    & \textbf{0.8156}    & \textbf{0.8854}    & \textbf{0.8087}    \\
\hline
\multicolumn{7}{c}{UCD}                                               \\
\hline
                 & Accuracy & \multicolumn{5}{c}{F1 score per class} \\
                 &          & Wake & S1   & S2   & SWS  & REM  \\
\hline
RF               & 0.5110        & 0.5328    & 0.0938    &  0.4129    &  0.7256    & 0.4098    \\
LSTM             & 0.5486        & 0.6232    & 0.1023    &  0.4624   & 0.8257    & 0.5158    \\
CNN              & 0.5650        &  0.6480    & 0.1450    &  0.4659    &  0.8274    & 0.5550    \\
GAT              & 0.5729        & 0.5141    & 0.3175     & 0.6006    & 0.6735    &  \underline{0.6938}    \\
MAML             & 0.6762        & 0.7148    & 0.3040    & 0.6877    &  \underline{0.8358}    & 0.6521    \\
MSL &  \underline{0.6894}        &  \underline{0.7695}    &  \underline{0.4740}    &   \underline{0.6935}    & 0.8065    & 0.6517    \\
MetaSTH-Sleep    & \textbf{0.7150}        & \textbf{0.7771}    & \textbf{0.5067}    & \textbf{0.7394}    & \textbf{0.8475}    &  \textbf{0.7079}    \\
\hline
\bottomrule
\end{tabular}
\end{table}
\textbf{Performance Comparison (RQ1):} We compare the overall performance of our proposed method with several baselines in terms of both overall accuracy and per-class F1 scores on two dataset. The best results are reported in Table~\ref{Tab: 1}. For ISRUC dataset, traditional machine learning models such as RF yield limited performance, achieveing an overall accuracy of only 0.5536 and struggling particularly with N1 (F1 = 0.1367), reflecting its inability to capture the temporal and spatial heterogeneity of EEG signals. Conventional deep learning models provides modest improvements: LSTM achieves 0.5253 accuracy, while CNN improves to 0.6183, benefiting from its spatial feature extraction ability. However, both models exhibit weak discriminability on sleep states, suggesting their limited adaptability to new subjects under limited data conditions. Graph-based learning with GAT further enhances performance with accuracy 0.5880, demonstrating improved spatial representation. Nonetheless, its F1 score for N3 indicates instability in modeling imbalanced classes. Meta-learning approaches, by contrast, show strong advantages in generalization. MAML and MSL provide better improvements on each classes, but it remains constrained by its shallow feature encoding. The proposed model MetaSTH-Sleep further pushes the frontier, achieving the highest overall accuracy of 0.8052. It also delivers the best F1 scores in all sleep stage classes, especially in hard-to-distinguish stages such as N1 and REM. This highlights the effectiveness of integrating spatial-temporal hypergraph with meta-learning, allowing the model to better capture dynamic interdependencies and adapt rapidly to new subjects under limited data conditions.
\begin{figure}[t]
    \centering
    \begin{subfigure}{0.32\textwidth}
        \includegraphics[width=\linewidth]{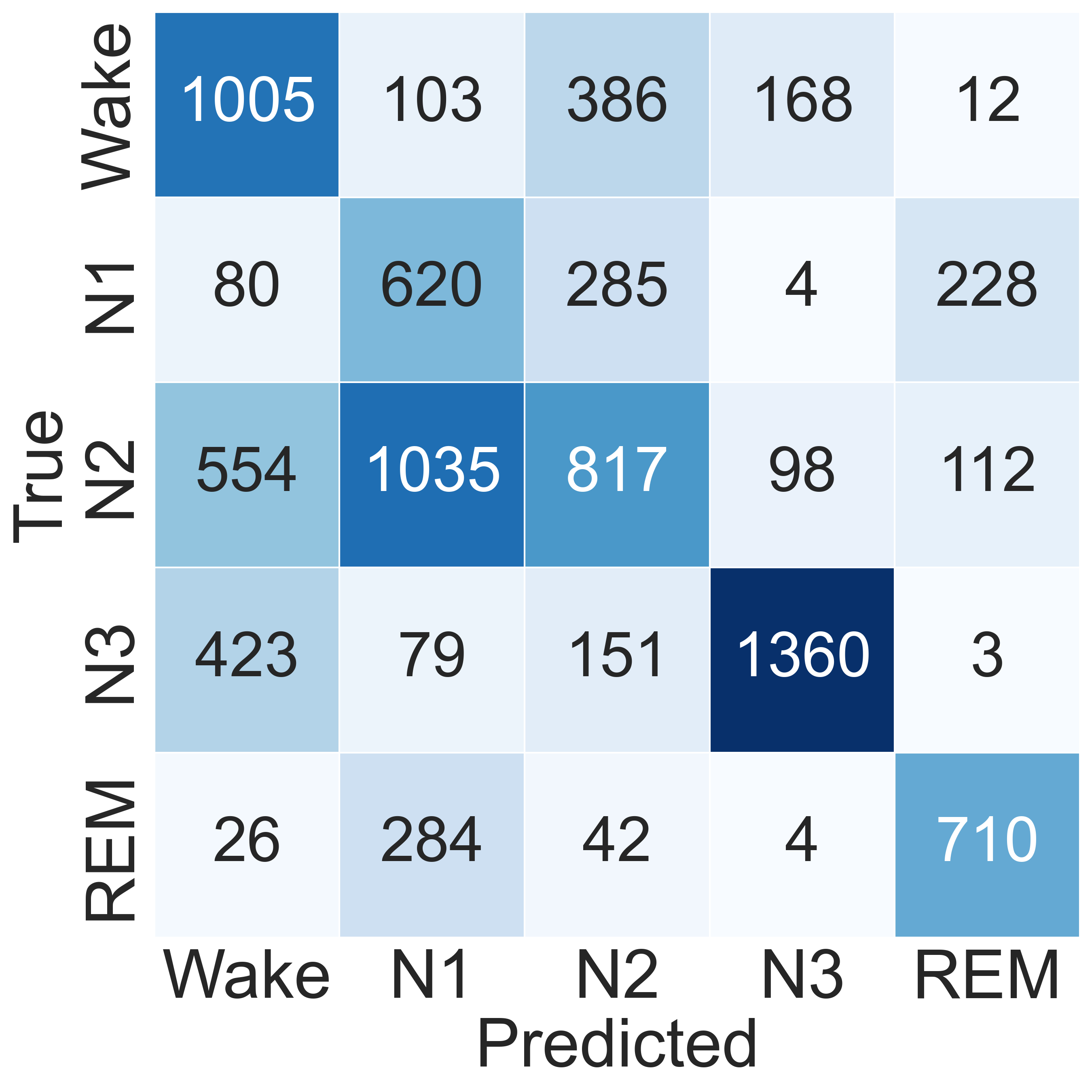}
        \caption{LSTM}
    \end{subfigure}
    \begin{subfigure}{0.32\textwidth}
        \includegraphics[width=\linewidth]{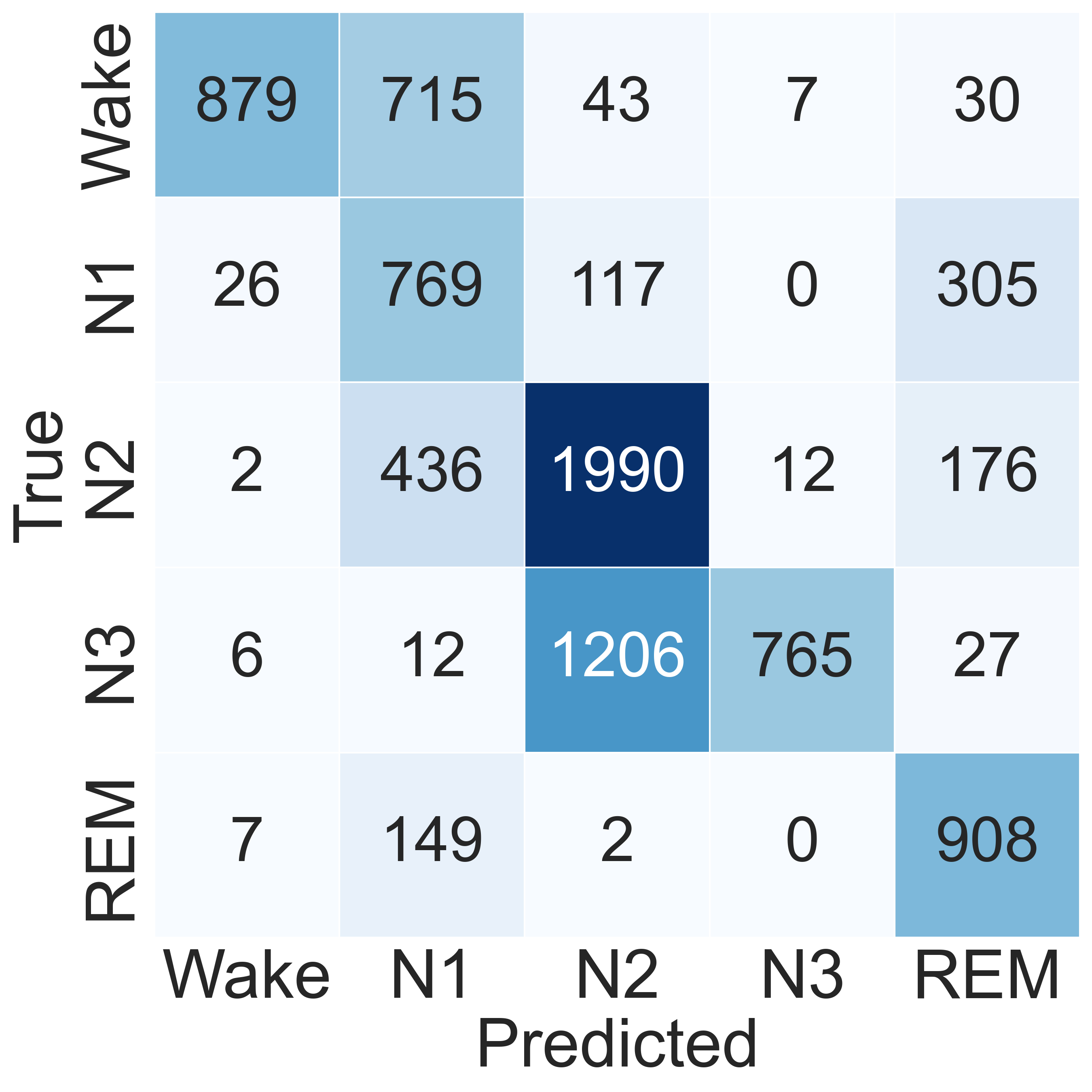}
        \caption{CNN}
    \end{subfigure}
    \begin{subfigure}{0.32\textwidth}
        \includegraphics[width=\linewidth]{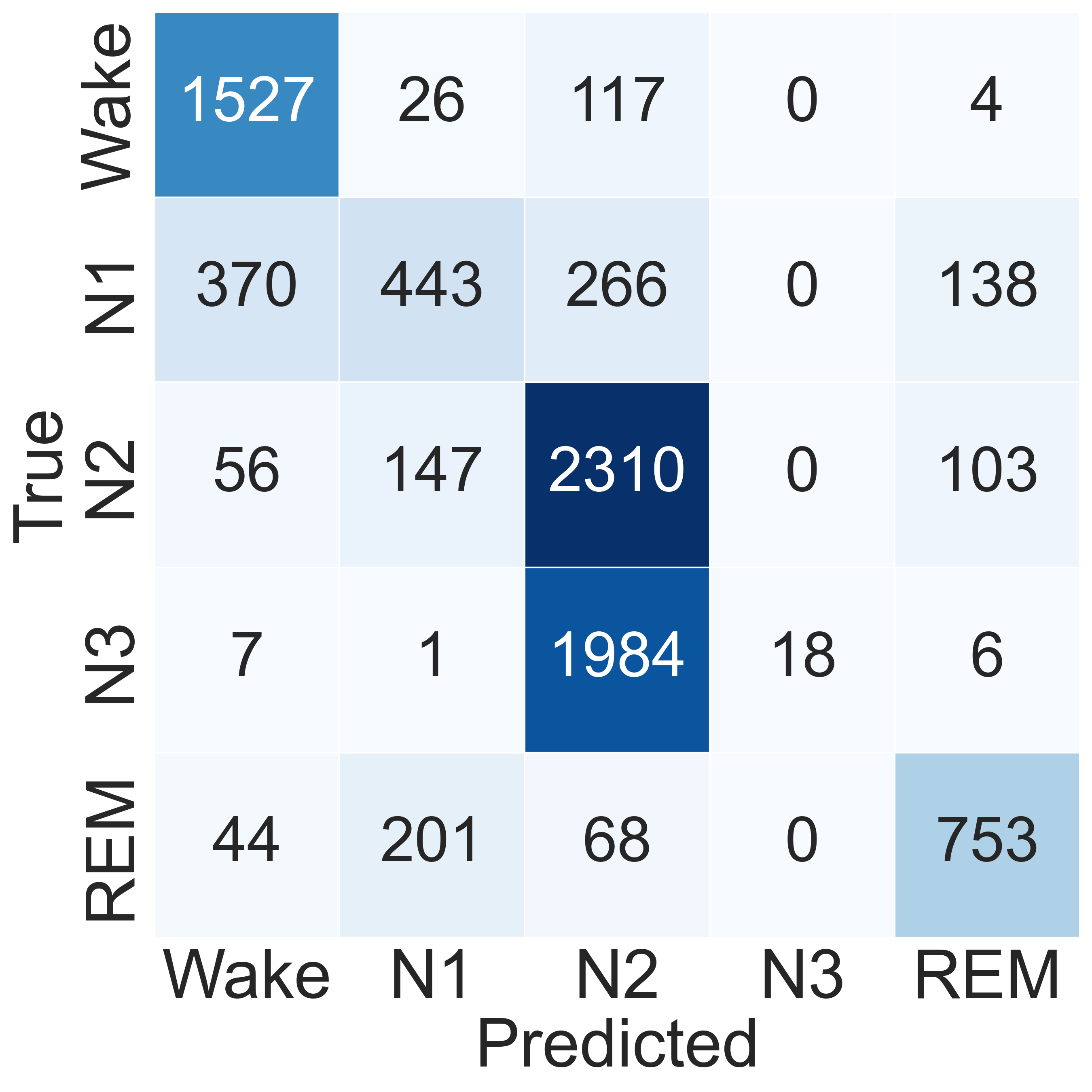}
        \caption{GAT}
    \end{subfigure}

    \vspace{0.3cm} % 可选：增加行间距

    \begin{subfigure}{0.32\textwidth}
        \includegraphics[width=\linewidth]{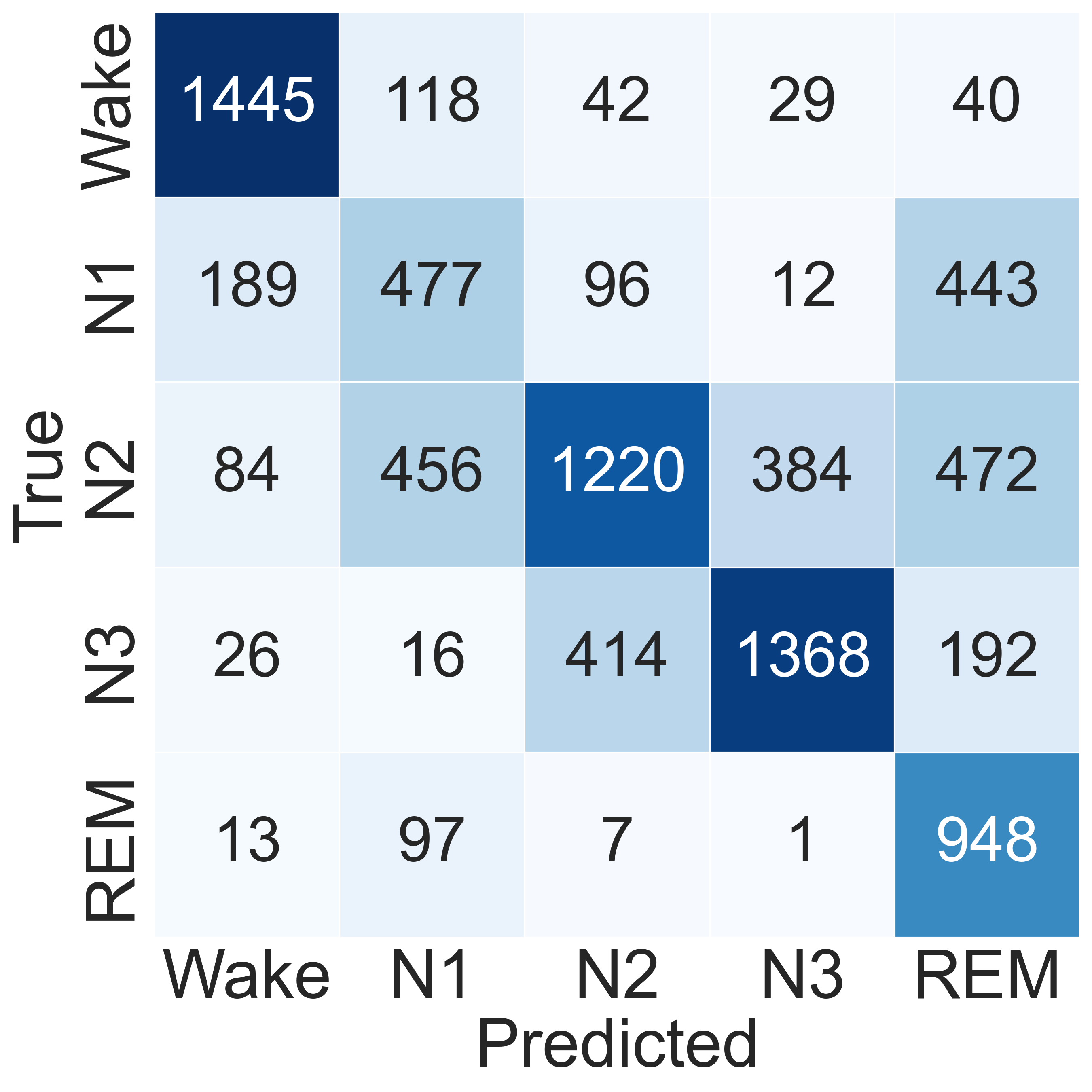}
        \caption{MAML}
    \end{subfigure}
    \begin{subfigure}{0.32\textwidth}
        \includegraphics[width=\linewidth]{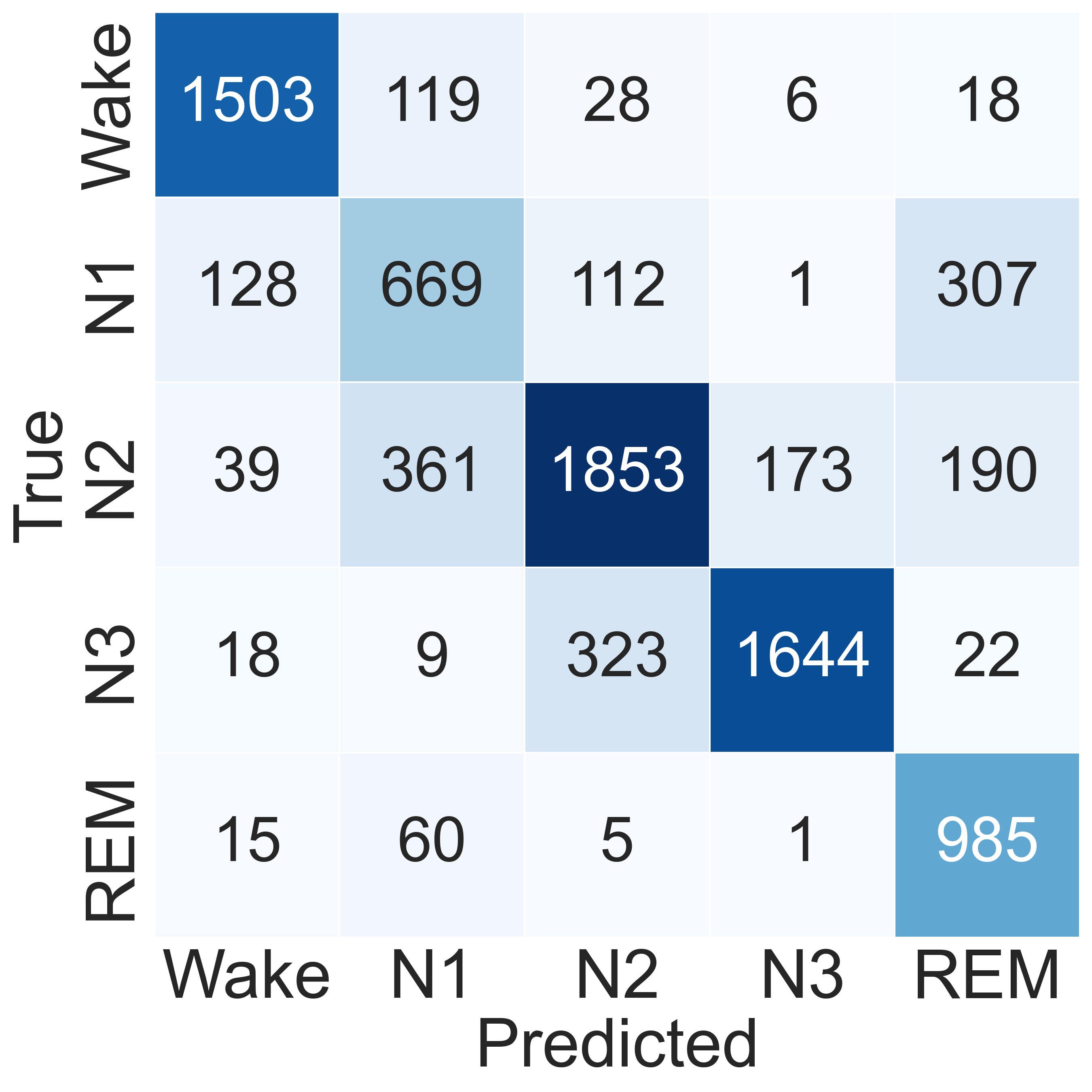}
        \caption{MSL}
    \end{subfigure}
    \begin{subfigure}{0.32\textwidth}
        \includegraphics[width=\linewidth]{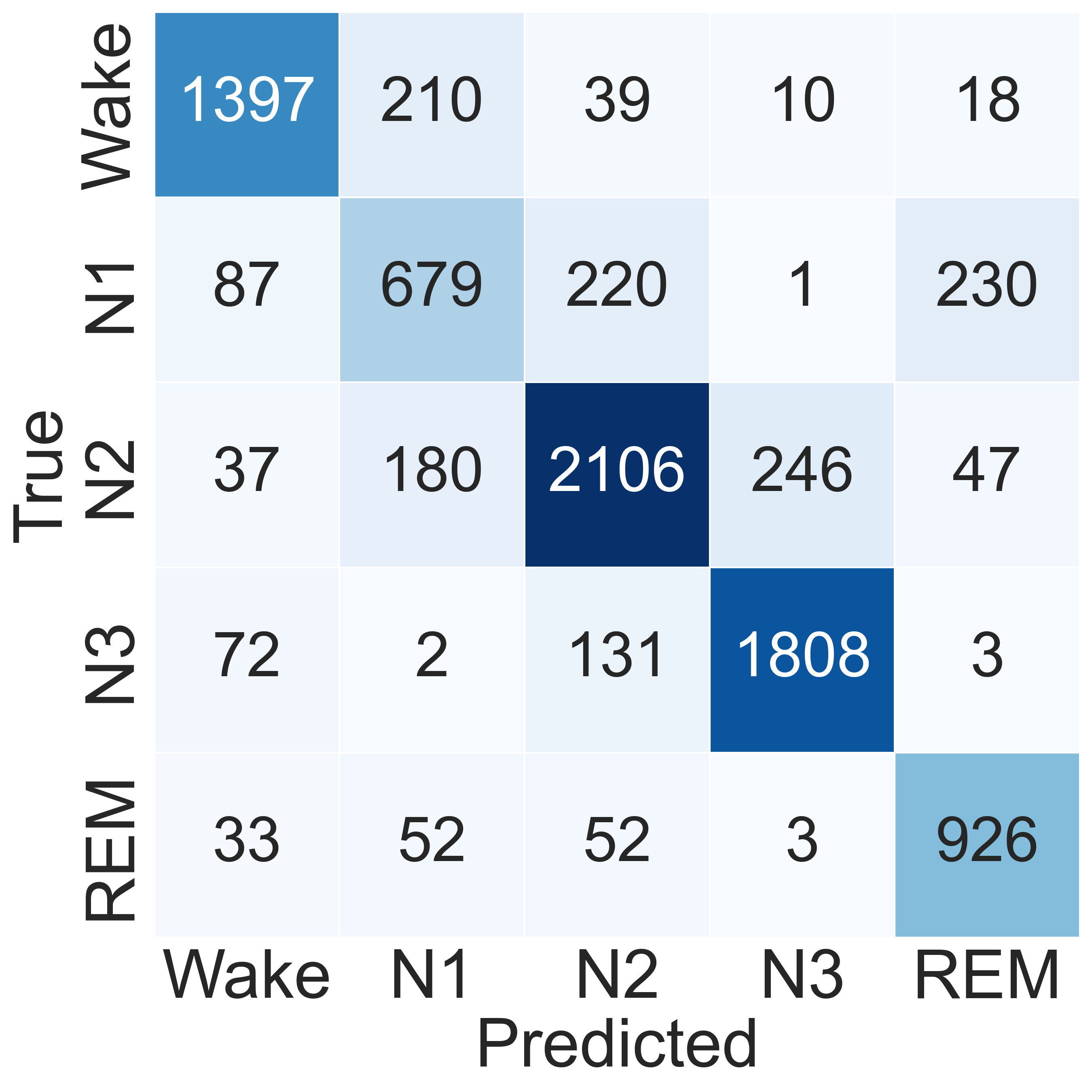}
        \caption{MetaSTH-Sleep}
    \end{subfigure}

    \caption{Confusion matrices of all compared methods on the ISRUC dataset.}
    \label{Fig:Confusion_Matrix}
\end{figure}
\\
\textbf{Confusion Pattern Analysis (RQ2):} To gain deeper insight into the strengths and limitations of each method in classifying different sleep stages, we analyzed their confusion matrices as shown in Fig.~\ref{Fig:Confusion_Matrix}.  These Matrices reveal not only overall predictive capability but also inter-class confusion patterns that are critical for evaluating clinical reliability, especially for sleep stages classification. Conventional deep learning models, such as CNN and LSTM, exhibit considerable confusion between between similar stages. For instance, the CNN model frequently misclassifies N2 and N3 into each other. LSTM, although capturing some temporal features, still suffers from severe confusion, particularly misclassifying Wake as N1 and REM as N1, reflecting its limited capacity in resolving temporal representation. As a graph-based model, GAT improves Wake and N2 classification but severely misclassifies N3 into N2, suggesting that spatial attention alone is insufficient for capturing deep sleep dynamics without temporal modeling. Moreover, REM and N1 remain highly confused, indicating its weakness in handling data limitation issue. MAML demonstrates better generalization across most stages, especially in N3 accuracy. However, it still suffers from moderate confusion between N2 and REM, and N1 misclassification persists due to its shallow feature encoder and limited spatial modeling. MSL mitigates confusion in most stages, especially between N2 and N3. Nevertheless, minor confusion still exists between REM and N1. Our proposed MetaSTH-Sleep shows the most distinguishable confusion matrix among all models. It achieves the highest true positive counts across all stages and notably reduces inter-stage confusion. This performance reflects the model’s enhanced ability to capture both spatial correlations and temporal continuity via its spatial-temporal hypergraph representation and rapid task adaptation through meta-learning.  The strong performance on N1 and REM demonstrates the effectiveness of our approach under few-shot and new subjects conditions.
\begin{table*}[t]
\centering
\caption{Performance comparison on different methods across various subjects on ISRUC. (*The best results are highlighted in bold; the runner up is underlined.)}
\label{Tab:all_subjects}
\small
\begin{tabularx}{\textwidth}{cc|XXXXXXX}
\toprule
\hline
Subject & Metric & RF & LSTM & CNN & GAT & MAML & MSL & \textbf{Ours} \\
\midrule
\multirow{2}{*}{1} & Accuracy  & 0.5536 & 0.5253 & 0.6183 & 0.5880 & 0.6355 & \underline{0.7773} & \textbf{0.8136} \\
                   & F1 score  & 0.5195 & 0.5394 & 0.6159 & 0.5277 & 0.6232 & \underline{0.7651} & \textbf{0.8051} \\ 
\multirow{2}{*}{2} & Accuracy  & 0.5104 & 0.4846 & 0.6185 & 0.5015 & 0.5686 & \underline{0.7627} & \textbf{0.8022} \\
                   & F1 score  & 0.5009 & 0.4962 & 0.5688 & 0.3890 & 0.5582 & \underline{0.7563} & \textbf{0.7983} \\
\multirow{2}{*}{3} & Accuracy  & 0.3282 & 0.6948 & 0.6159 & 0.5763 & 0.6407 & \underline{0.7865} & \textbf{0.8255} \\
                   & F1 score  & 0.3365 & 0.6594 & 0.6256 & 0.4976 & 0.6317 & \underline{0.7783} & \textbf{0.8187} \\
\multirow{2}{*}{4} & Accuracy  & 0.3843 & 0.5984 & 0.5565 & 0.5579 & 0.6026 & \underline{0.7694} & \textbf{0.7995} \\
                   & F1 score  & 0.3174 & 0.5855 & 0.5478 & 0.4571 & 0.5916 & \underline{0.7608} & \textbf{0.7878} \\
\multirow{2}{*}{5} & Accuracy  & 0.4759 & 0.7012 & 0.6386 & 0.5998 & 0.6387& \underline{0.7744} & \textbf{0.8086} \\
                   & F1 score  & 0.4673 & 0.6986 & 0.6595 & 0.5516 & 0.6280 & \underline{0.7677} & \textbf{0.8008} \\ \hline
\multirow{2}{*}{6} & Accuracy  & 0.5354 & 0.6274 & 0.6259 & 0.6316 & 0.6258 & \underline{0.7853} & \textbf{0.8037} \\
                   & F1 score  & 0.4795 & 0.6088 & 0.6145 & 0.5746 & 0.6130 & \underline{0.7774} & \textbf{0.7975} \\
\multirow{2}{*}{7} & Accuracy  & 0.4694 & 0.7435 & 0.5865 & 0.5968 & 0.6469 & \underline{0.7684} & \textbf{0.8068}\\
                   & F1 score  & 0.4775 & 0.7308 & 0.6024 & 0.5418 & 0.6345 & \underline{0.7575} & \textbf{0.7944} \\
\multirow{2}{*}{8} & Accuracy  & 0.4633 & 0.6876 & 0.5489 & 0.7662 & 0.7012 & \underline{0.7893} & \textbf{0.8293} \\
                   & F1 score  & 0.4780 & 0.6809 & 0.5245 & 0.7463 & 0.6883 & \underline{0.7779} & \textbf{0.8183} \\
\multirow{2}{*}{9} & Accuracy  & 0.5026 & 0.5991 & 0.5375 & 0.6015 & 0.6691 & \underline{0.7816} & \textbf{0.8091} \\
                   & F1 score  & 0.4529 & 0.6119 & 0.5206 & 0.5499 & 0.6567 & \underline{0.7688} & \textbf{0.7989}\\
\multirow{2}{*}{10} & Accuracy  & 0.3592 & 0.6110 & 0.5408 & 0.6638 & 0.6764 & \underline{0.7419} & \textbf{0.7755} \\
                   & F1 score  & 0.3690 & 0.6078 & 0.5396 & 0.6392 & 0.6657 & \underline{0.7348} & \textbf{0.7687} \\
\hline
\bottomrule
\end{tabularx}
\end{table*}
\\
\textbf{Robustness Analysis (RQ3):} To evaluate the robustness of each method under inter-subject variability, we conducted subject-wise testing, treating each of the 10 subjects as meta-test target while training on the remaining ones. Table~\ref{Tab:all_subjects} shows the classification accuracy and F1 score for all compared methods across unseen individual subjects. The results indicate that traditional machine learning methods such as RF exhibit large performance fluctuations across subjects, with accuracy ranging from 0.3282 to 0.5536 and F1 scores as low as 0.3174. This inconsistency reflects the model unable to generalize across individuals, due to its reliance on lack of temporal modeling capabilities. Conventional deep learning models, including LSTM and CNN, show moderate improvements but still suffer from significant instability. LSTM performs well on certain subject (e.g., Subject 7), but fails on others (e.g., Subject 2). This indicating that sequence modeling alone is insufficient for robust unseen subject generalization. Graph-based model, such as GAT further improve spatial representation learning, achieving relatively high F1 scores on subject like Subject 8. However, their robustness remains limited due to a lack of temporal modeling, leading to weak performance on Subject 2 and Subject 3. In contrast, meta-learning models demonstrate significantly higher robustness across all subjects. MAML and MSL improves consistency across individuals but still exhibits fluctuations due to its shallow feature encoder and sensitivity to imbalance tasks during meta-training. Our proposed MetaSTH-Sleep outperforms all baselines on every subject, achieving the highest accuracy and F1 scores. For example, it obtains 0.8293 accuracy and 0.8183 F1 score on Subject 8, and maintains competitive scores even on more challenging subjects like Subject 10. These results validate the effectiveness of (1) spatial-temporal hypergraph allowing the model to capture the higher-order relations existed in both spatial and temporal dimension, and (2) the meta-learning framework facilitates fast adaptation to new subjects using only a few labeled samples.
\begin{figure}[t]
    \centering
    \begin{subfigure}{0.4\textwidth}
    \includegraphics[width=\linewidth]{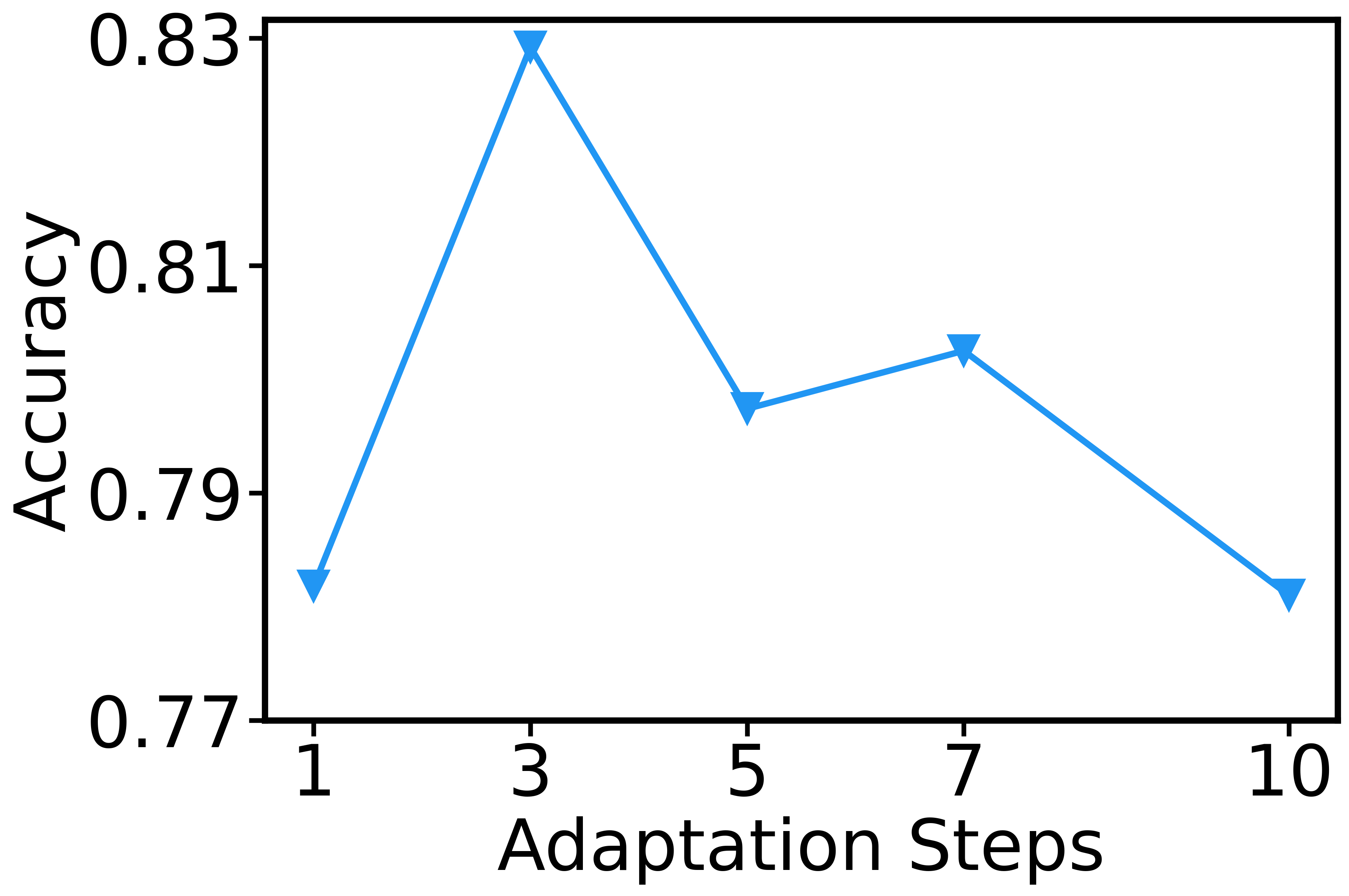}
    \caption{Accuracy}
    \label{Fig:adaption_acc}
    \end{subfigure}
    \begin{subfigure}{0.4\textwidth}
    \includegraphics[width=\linewidth]{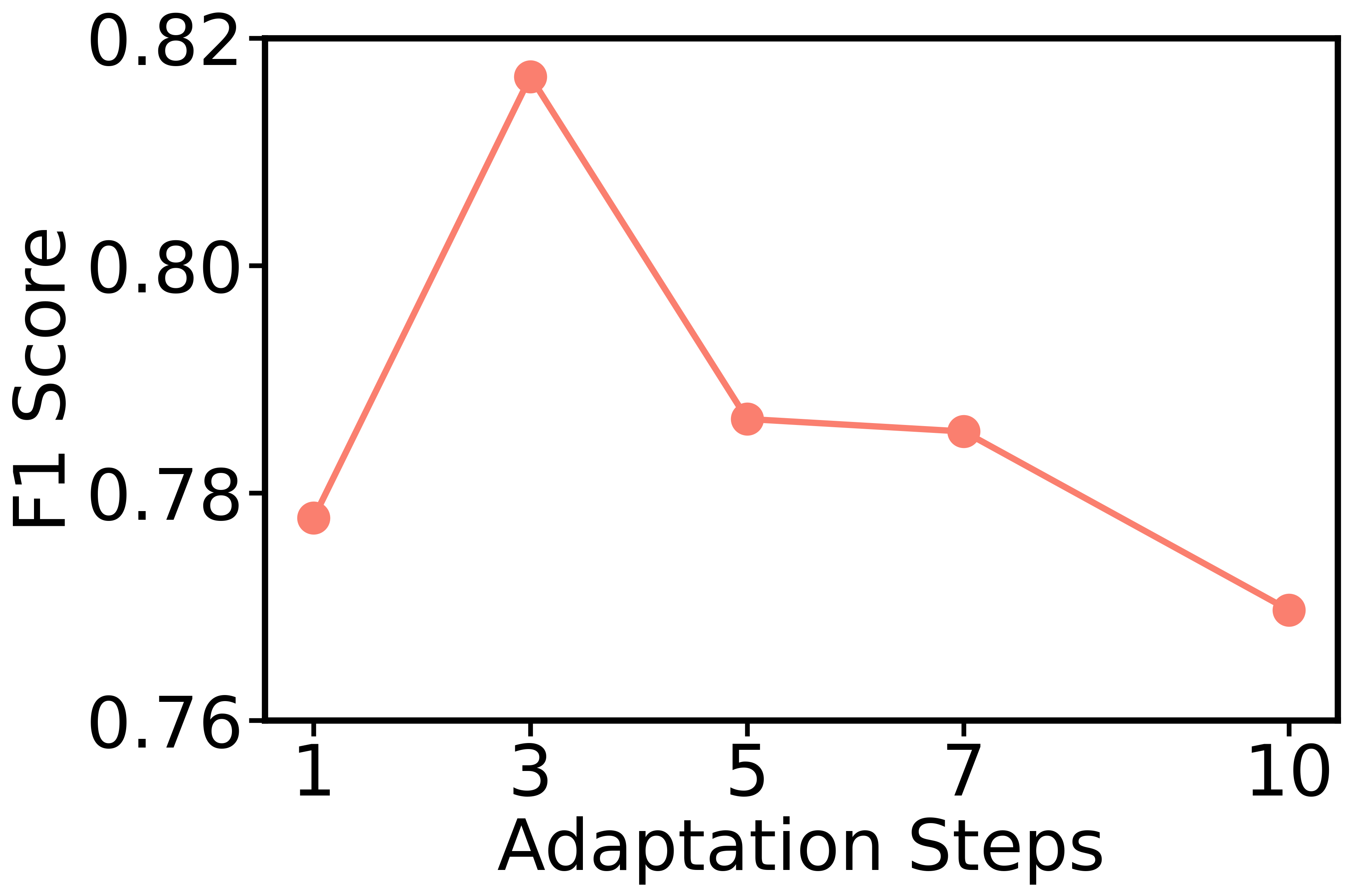}
    \caption{F1 Score}
    \label{Fig:adaption_f1}
    \end{subfigure}   
    \caption{Performance of MetaSTH-Sleep under varying adaptation steps.}
    \label{Fig:adaption}
\end{figure}
\begin{figure}[t]
    \centering
    \begin{subfigure}{0.41\textwidth}
    \includegraphics[width=\linewidth]{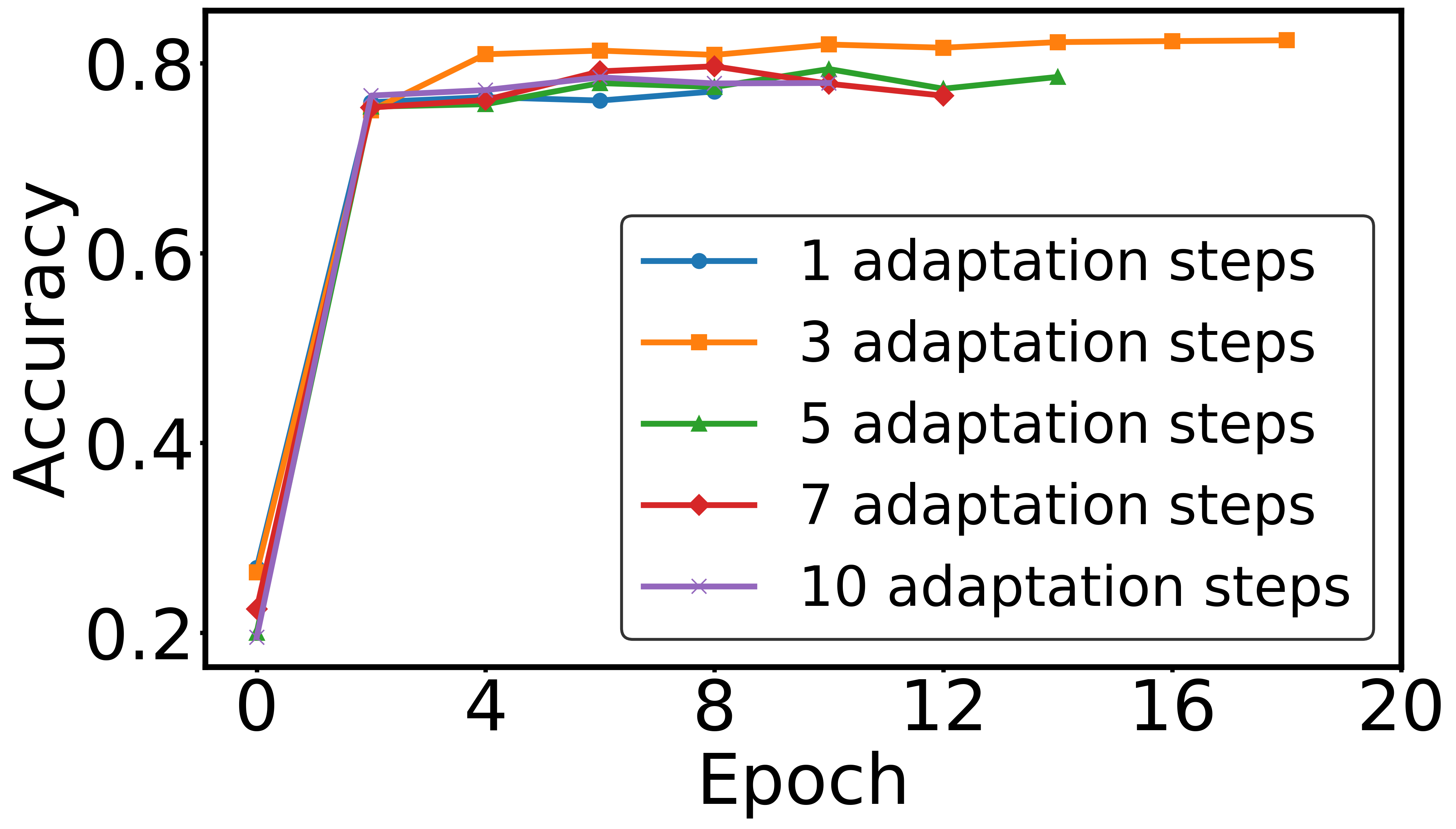}
    \caption{Accuracy}
    \label{Fig:adaption_epoch_acc}
    \end{subfigure}
    \begin{subfigure}{0.4\textwidth}
    \includegraphics[width=\linewidth]{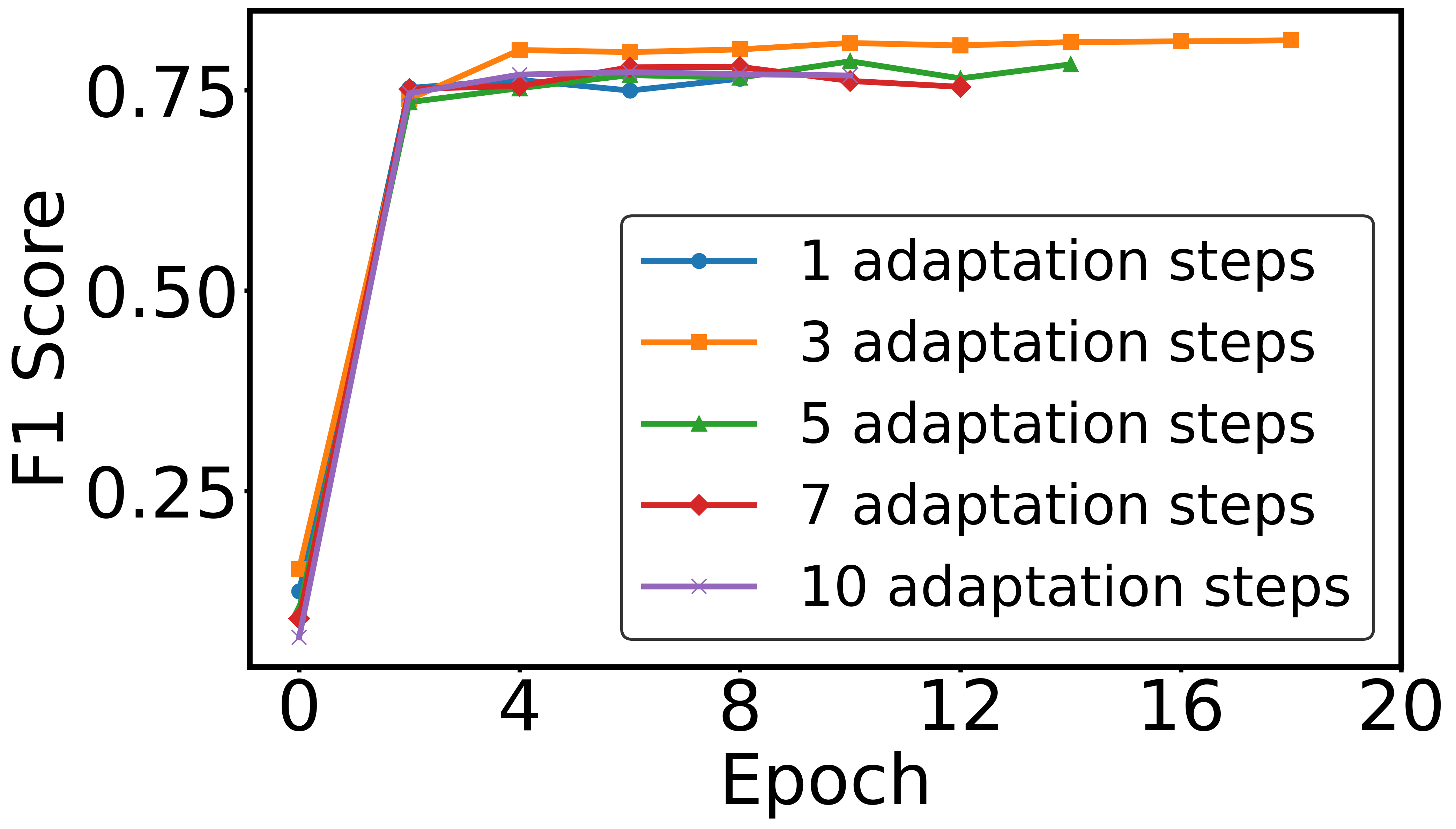}
    \caption{F1 score}
    \label{Fig:adaption_epoch_f1}
    \end{subfigure}   
    \caption{Convergence performance of MetaSTH-Sleep across different adaptation steps.}
    \label{Fig:adaption_epoch}
\end{figure}
\\
\textbf{Impact of Adaptation Steps and Support Set Size(RQ4):} To evaluate the fast adaptation capability of our model, we analyzed its performance under varying numbers of adaptation steps. As shown in Fig.~\ref{Fig:adaption_acc} and Fig.~\ref{Fig:adaption_f1}, the performance initially increases and then fluctuates with more adaptation steps. Specifically, accuracy peaks at 3 adaptation steps and then slightly declines with further increases, indicating a potential risk of overfitting to limited data. A similar trend is observed in the F1 score, which reaches its maximum at 3 steps before decreasing, suggesting that a moderate number of steps allows for optimal generalization.

In addition, Fig.~\ref{Fig:adaption_epoch_acc} and Fig.~\ref{Fig:adaption_epoch_f1} further highlights the efficiency of 3 adaptation steps in achieving rapid convergence. Specifically, the model with 3 steps reaches its peak validation accuracy and F1 score within fewer epochs compared to other settings, demonstrating a balance between adaptation speed and generalization. This suggests that for the given dataset, 3 adaptation steps allow meta-learning to effectively learn task-specific features. In contract, fewer steps slow down convergence, while more steps introduce more noise. These findings underscore the adaptation steps in optimizing meta-learning for few-shot sleep stage classification.
\begin{figure}[t]
    \centering
    \begin{subfigure}{0.4\textwidth}
    \includegraphics[width=\linewidth]{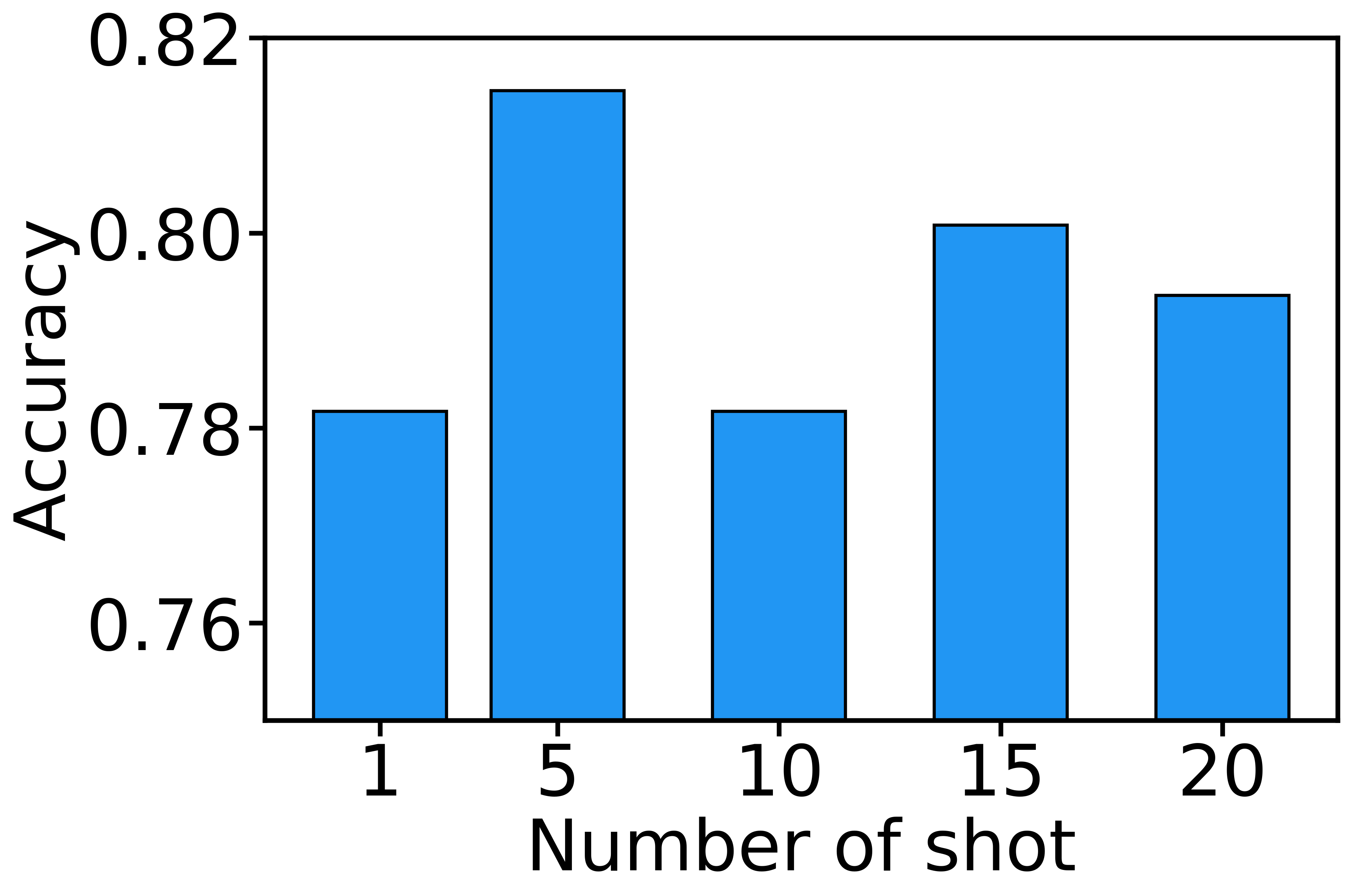}
    \caption{Accuracy}
    \label{Fig:nshot_accuracy}
    \end{subfigure}
    \begin{subfigure}{0.4\textwidth}
    \includegraphics[width=\linewidth]{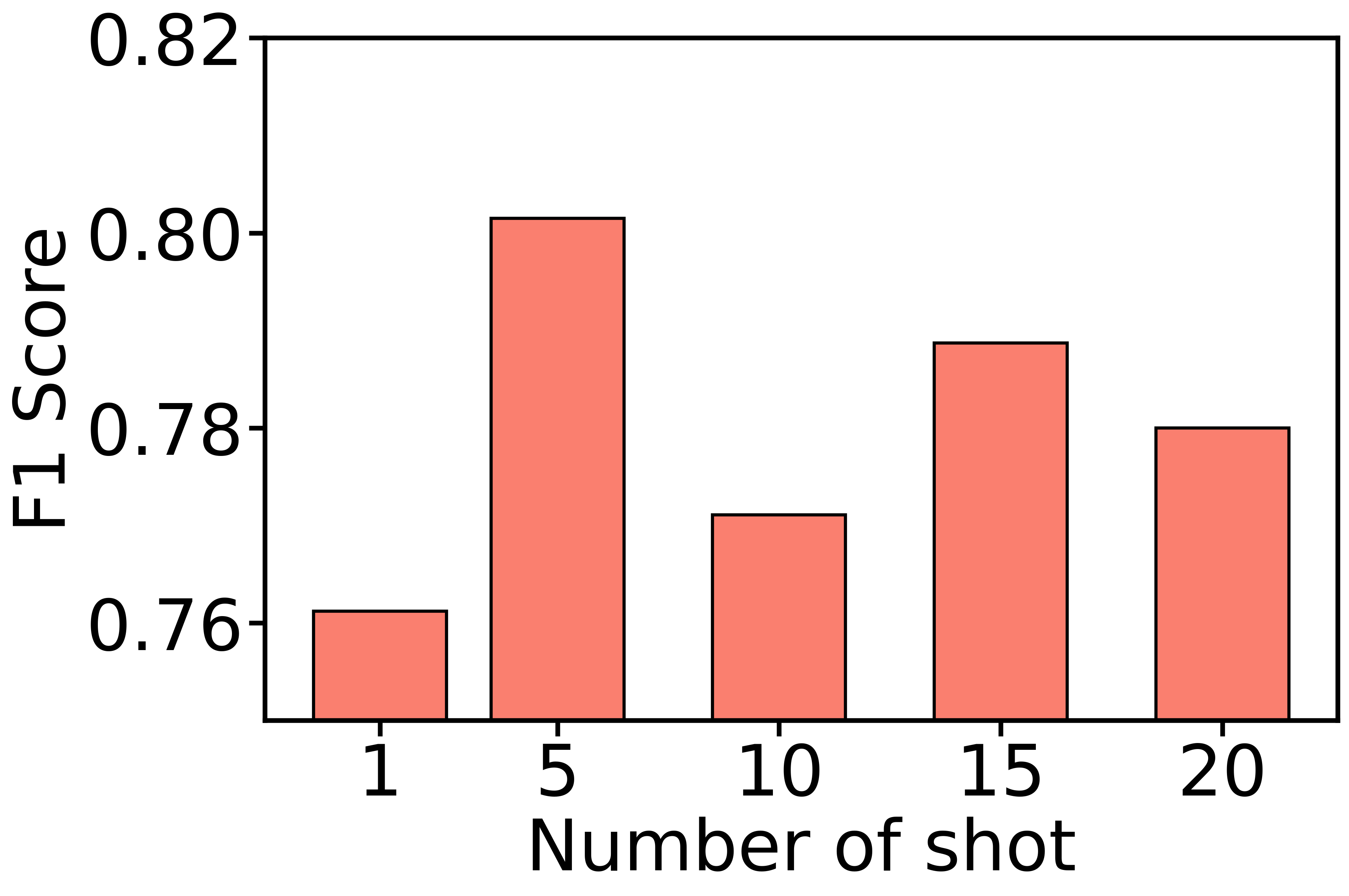}
    \caption{F1 score.}
    \label{Fig:nshot_f1score}
    \end{subfigure}   
    \caption{Few-shot performance of MetaSTH-Sleep under different support set sizes.}
    \label{Fig:}
\end{figure}

We further evaluated the the number of support size per class (i.e., n-shot) to investigate its few-shot learning performance under limited data scenarios. As shown in Fig.~\ref{Fig:nshot_accuracy} and Fig.~\ref{Fig:nshot_f1score}, the model exhibits its highest performance at 5-shot, achieving the peak accuracy and F1 score of 0.8136 and 0.8052, respectively. Notably, both metrics drop when increasing the number of shots afterwards. This trend suggests that although larger support size generally provide more data, excessive samples may introduce label noise, reduce task consistency. Meanwhile, performance at 1-shot is also insufficient learning, indicating that a minimal support sets lacks information for effective adaptation. These findings suggest that there exists an optimal range of support set size that balances the trade-off between information sufficiency and task overfitting. For MetaSTH-Sleep, the
best setting is 5 for all subjects.

\begin{table}[t]
\centering
\caption{Performance of accuracy for MetaSTH-Sleep ablation study.}
\label{tab:ablation_accuracy}
\small
\begin{tabular}{l|cc}
\toprule
\toprule
\textbf{Model} & \textbf{ISRUC} & \textbf{UCD} \\
\midrule
MetaSTH-Sleep & 0.8052 & 0.7150 \\
-w/o Hypergraph Construction & 0.7691 & 0.6856 \\
-w/o Multi-head Attention & 0.6383 & 0.6684 \\
\bottomrule
\bottomrule
\end{tabular}
\end{table}

\textbf{Ablation Study (RQ5):} 
In order to verify the effectiveness of different components in MetaSTH-Sleep, two variants are constructed and compared with the complete model on the sleep stage classification task. The results are presented in Table~\ref{tab:ablation_accuracy}. In particular, the effects of spatial-temporal hypergraph construction and multi-head attention are examined. The variant \textit{w/o Hypergraph Construction} replaces the reconstruction-based process with simple pairwise connections, where edges are directly formed between nodes without generating high-order hyperedges. MetaSTH-Sleep consistently outperforms this variant, with accuracy improvements of about 3–4\% on both datasets, indicating that the hypergraph formulation better preserves complex spatial-temporal dependencies. The variant \textit{w/o Multi-head Attention} removes the adaptive weighting mechanism and treats spatial and temporal hyperedges equally. This leads to a marked decrease in accuracy, especially on ISRUC over 16\%, which demonstrates the necessity the multi-head attention mechanism.

% \textcolor{red}{
% 1. Performance Analysis - Table 1-
% We evaluated the accuracy and per-class F1 score of our method on two datasets to assess overall classification performance.\\
% 2. Performance Analysis - Figure- (Confusion Matrix) We visualized the confusion matrices of four different methods to compare their classification error patterns. \\
% 3. Robustness- Table 2- We evaluated all methods by treating each of 10 subjects as the meta-test subject. \\
% 4. (Adaptation Steps) - Figure-We analyzed performance variation under different numbers of adaptation steps to evaluate fast adaptation capability. \\
% 5.(Support Set Size) - Figure - We analyzed performance across varying support set sizes (n-shot) to demonstrate few-shot learning under limit data scenario.
% }

\section{Conclusion}
In this paper, we propose a few-shot sleep stage classification framework based on spatial-temporal hypergraph enhanced meta-learning, namely MetaSTH-Sleep. The framework facilitates rapid and efficient adaptation to new subjects under limited data conditions, effectively addressing the challenge of data scarcity. Most importantly, by incorporating a spatial-temporal hypergraph into the meta-learning paradigm, The framework is capable of capturing the hidden spatial correlations and temporal dependencies from the EEG waves simultaneously, leading to powerful representation learning from EEG signals. The effectiveness and superiority of MetaSTH-Sleep are validated through comprehensive experiments against a variety of baselines and model variants. Furthermore, extensive sensitivity analyses demonstrate the robustness of the model with respect to key meta-learning parameters such as the number of adaptation steps and the support set size. The experimental results confirm that the proposed framework offers a generalizable and effective solution for leveraging spatial-temporal hypergraphs and meta-learning in few-shot sleep stage classification scenarios,\textcolor{black}{ which in turn contributes to more reliable health management of sleep and related disorders.}

\bibliographystyle{elsarticle-num} 
\bibliography{bibliography}

\end{document}